\documentclass[10pt,journal,compsoc]{IEEEtran}
\usepackage{cite}
\usepackage{graphicx}
\usepackage{epstopdf}
\usepackage{bbm}
\usepackage[linesnumbered,titlenumbered,ruled,vlined,commentsnumbered,noend]{algorithm2e}

\usepackage[tight]{subfigure}
\usepackage{multirow}
\usepackage{xcolor}
\usepackage{array}
\usepackage{caption}
\usepackage{rotating}
\usepackage{verbatim}
\usepackage{amsmath}
\begin{document}
	\title{Local Black-box Adversarial Attacks: A Query Efficient Approach}

    \author{Tao Xiang,~\IEEEmembership{Member,~IEEE,}
        Hangcheng Liu,
        Shangwei Guo,
        Tianwei Zhang,
        and Xiaofeng Liao,~\IEEEmembership{Fellow,~IEEE}% <-this % stops a space
        \IEEEcompsocitemizethanks{
        \IEEEcompsocthanksitem \textcolor{red}{\textbf{This work has been submitted to the IEEE for possible publication. Copyright may be transferred without notice, after which this version may no longer be accessible.}}
        \IEEEcompsocthanksitem T. Xiang, H. Liu, S. Guo, and X. Liao are with the College of Computer Science, Chongqing University, Chongqing 400044, China. E-mail: \{txiang; hcliu; swguo; xfliao\}@cqu.edu.cn.
        \IEEEcompsocthanksitem T. Zhang is with the School of Computer Science and Engineering, Nanyang Technological University, Singapore. E-mail: tianwei.zhang@ntu.edu.sg.}
    }

	\maketitle
	
	\begin{abstract}	
	Adversarial attacks have threatened the application of deep neural networks in security-sensitive scenarios. Most existing black-box attacks fool the target model by interacting with it many times and producing global perturbations. However, global perturbations change the smooth and insignificant background, which not only makes the perturbation more easily be perceived but also increases the query overhead.

	In this paper, we propose a novel framework to perturb the discriminative areas of clean examples only within limited queries in black-box attacks. Our framework is constructed based on two types of transferability. The first one is the transferability of model interpretations. Based on this property, we identify the discriminative areas of a given clean example easily for local perturbations. The second is the transferability of adversarial examples. It helps us to produce a local pre-perturbation for improving query efficiency. After identifying the discriminative areas and pre-perturbing, we generate the final adversarial examples from the pre-perturbed example by querying the targeted model with two kinds of black-box attack techniques, i.e., gradient estimation and random search. We conduct extensive experiments to show that our framework can significantly improve the query efficiency during black-box perturbing with a high attack success rate. Experimental results show that our attacks outperform state-of-the-art black-box attacks under various system settings.
   \end{abstract}

	
	\IEEEpeerreviewmaketitle	
	\section{Introduction} \label{sec:introduction}
Deep Neural Networks (DNNs) are easily fooled by adversarial examples \cite{szegedy2014} that are generated by adding some tiny and carefully crafted invisible perturbations to clean examples. Such malicious examples can mislead Deep Learning (DL) models into making wrong predictions without being perceived by human beings. Due to the critical threat, adversarial examples have been becoming a great challenge when applying deep learning in security-sensitive scenarios \cite{bose2018,xie2017,carlini2018audio,papernot2016,wang2019security,sun2020counteracting} such as autonomous driving \cite{sun2020counteracting}.

Compared with white-box adversarial example generation attacks \cite{szegedy2014, goodfellow2015explaining,papernot2016,moosavi2016deepfool,carlini2017towards,Moosavi2017Universal} that allow adversaries to access the architecture and parameters of the target model, black-box attacks \cite{brendel2018decision, narodytska2017simple, bhagoji2018practical,ilyas18a,papernot2017practical,tu2019autozoom,Yanpei2017delving,suya2020hybrid,li2019adversarial} are more threatening and practical in real applications, where an adversary can only query the target model via application programming interfaces (APIs). Due to their serious threat, we, from an adversarial aspect, focus on black-box attacks in this paper.

\begin{figure}[t]
	\centering
	\subfigure[]{\includegraphics[width=0.3\columnwidth]{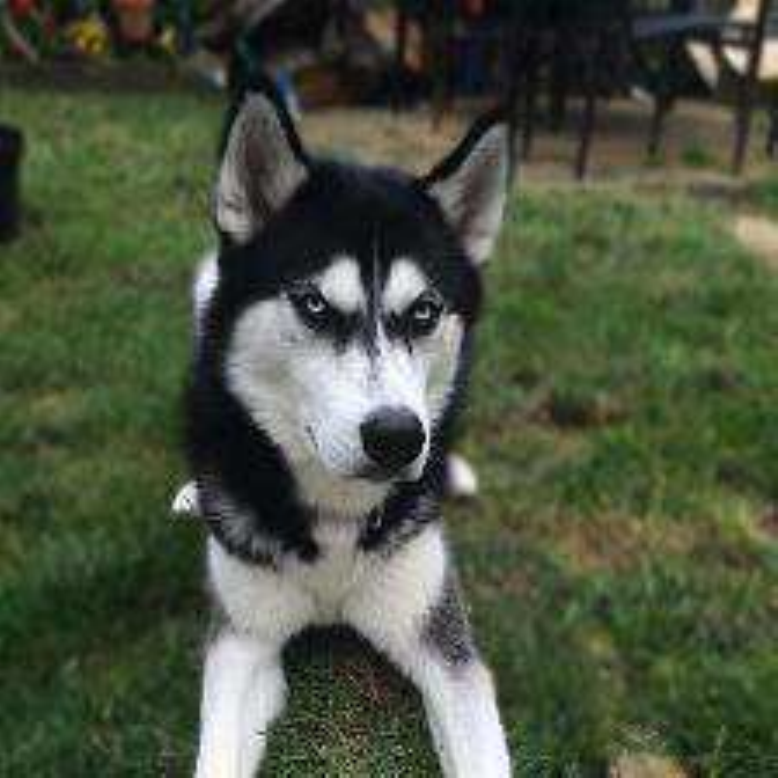}}
	\subfigure[]{\includegraphics[width=0.3\columnwidth]{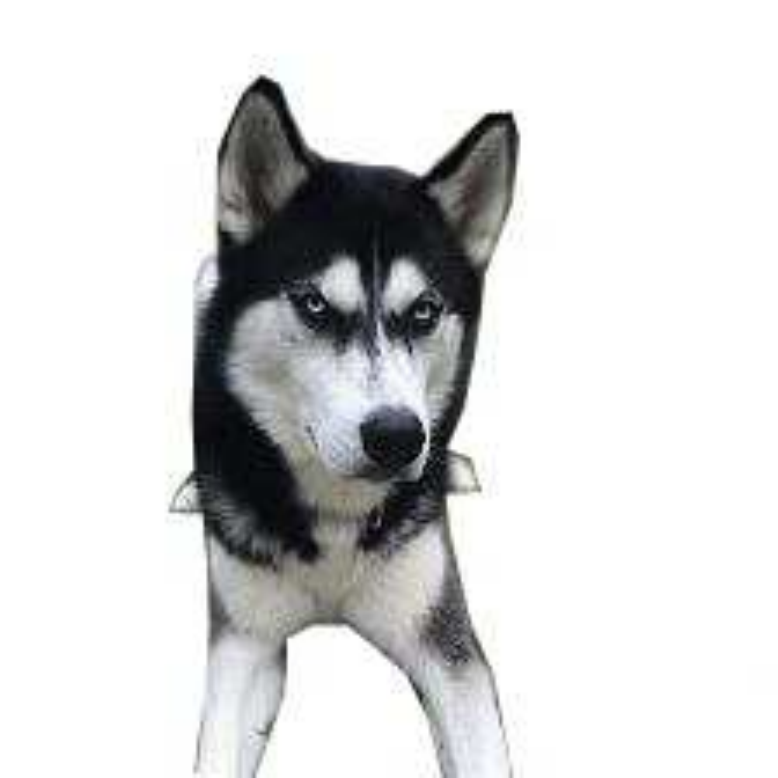}}
	\subfigure[]{\includegraphics[width=0.3\columnwidth]{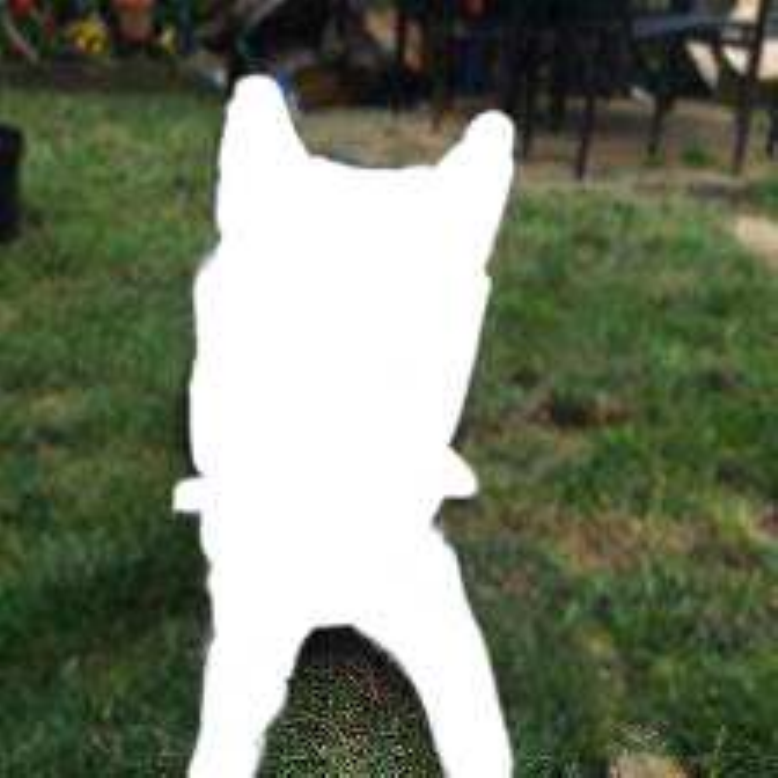}}	
	
	\caption{Not all pixels are born equal: (a) clean example that is rightly predicted as Siberian husky by ResNet50 with 51.82\% confidence; (b) example without background, which is still predicted as Siberian husky with 66.04\% confidence; (c) example with only background, which is wrong predicted as sulphur-crested cockatoo with 45.22\% confidence and the confidence of Siberian husky is only 0.26\%. }
	\label{fig:not_equal}
\end{figure}

Although quantities of black-box attacks have been proposed, most of them generate global perturbations for adversarial examples  \cite{brendel2018decision, narodytska2017simple, bhagoji2018practical,ilyas18a,papernot2017practical,tu2019autozoom,Yanpei2017delving,suya2020hybrid}, in which each pixel of a clean example may be perturbed. However, \emph{not all pixels are born equal}. Intuition tells us that the background of an image should not affect the classification result, just as shown in Fig. \ref{fig:not_equal}. To generate adversarial examples, a black-box attack also requires querying the targeted model many times, which is resource-consuming and may be detected. So we raise this question: \emph{is it possible for an adversary to generate invisible local perturbations to fool the targeted model, while only requiring a small number of queries?}

Roughly speaking, the generation of local perturbations consists of two steps. First, the adversary needs to identify the discriminative pixels/areas of a given clean example. Second, the adversary perturbs the discriminative pixels by querying the targeted model until the target model is fooled. Several works \cite{papernot2016,li2019adversarial,9157746} have proposed attacks to generate local perturbations. In particular, for the first step, they adopt salient object detection techniques to identify the discriminative areas. For example, Dong \textit{et al.} \cite{9157746} leverage the class activation mapping (CAM) \cite{zhou2016learning} to produce salience maps. Li \textit{et al.} \cite{li2019adversarial} utilize a pre-trained semantic segmentation model to divide a clean example into foreground and background.

Unfortunately, these salient object detection techniques either rely on the white-box access of the targeted model or consume a huge amount of computational resources. In particular, CAM requires the internal information of the targeted model and needs to change the architecture of the target model to produce discriminative areas, which cannot directly be applied to black-box scenarios. On the other hand, it is difficult to own a well-trained semantic segmentation model that is produced based on high-quality datasets and a huge amount of computational resources. For classification tasks, training a semantic segmentation model is even harder than the training of the targeted model. Thus, existing discriminative area identification methods are either inapplicable or inefficient for black-box attacks.

For the second step, one straightforward way is to directly adopt global perturbation techniques (e.g. gradient estimation \cite{chen2017zoo,ilyas2018black,bhagoji2018practical}  and random search \cite{li2019adversarial,narodytska2017simple,brendel2018decision}) for local perturbation as Li \textit{et al.} \cite{li2019adversarial} did. However, this approach still requires a huge amount of interactions with the targeted model to pass across the decision boundary because the start point is still a clean example. Besides, as we will show in Section \ref{sec:experiments}, the local perturbation in \cite{li2019adversarial} is easy to cause visible noises because they change each perturbed pixel as the same (e.g. (255,255,255)).

Inspired by the above challenges, we propose a novel black-box, query-efficiency framework to generate local perturbations in this paper. Our design strategies are guided by our two key observations. The first one is the weak transferability of adversarial examples: a perturbed example produced by an arbitrary trained model is closer to the decision boundary of the target model. Taking advantage of the weak transferability of adversarial examples, we propose to utilize a cheap or publicly available reference model for producing a pre-perturbation for a clean example such that the corresponding pre-perturbed example approaches the boundary of the target model. Thus, compared with the clean example, the pre-perturbed example can save numerous black-box queries.

The second observation is the transferability of model interpretations. We found that the model interpretations overlap greatly when interpreting different models using the same model interpreter. This property indicates that the discriminative areas of the clean example for the target model can be obtained through the same reference model for pre-perturbation and without any extra complex semantic segmentation model.

Our framework generates local perturbed adversarial examples by two phases: preprocessing and local black-box perturbing. In the first phase, taking advantage of the transferability of model interpretations, we utilize a cheap or publicly available reference model to identify discriminative areas of a given clean example. We also generate the pre-perturbed example through the same reference model based on the transferability of adversarial examples, which is closer to the decision boundary of the target model than the clean example. In the second phase, we perturb the pre-perturbed example in a black-box way instead of perturbing the clean example from scratch for much better query efficiency. Further, we implement the local black-box attacks based on gradient estimation and random search, respectively.

We conduct comprehensive experiments to show that the proposed attacks successfully avoid perturbing the smoother background and bring better visual effect. Besides, our attacks also achieve 1.68X-5.16X performance improvement of query efficiency over existing state-of-the-art black-box attacks. 

Our main contributions can be summarized as follows:
\begin{itemize}
	\item We find and utilize the transferability of model interpretations to identify the discriminative areas in black-box adversarial attacks.
	\item We pre-perturb discriminative pixels before back-box interaction, which can significantly reduce the number of queries to the targeted model.
	\item We propose a universal query efficient attack framework to generate local adversarial perturbations that can be applied to different types of global black-box attacks.
	\item We conduct extensive experiments to illustrate the effectiveness and efficiency of our attack under various system settings.
\end{itemize}

The rest of this paper is organized as follows. Section \ref{sec:relatedwork} summarizes existing state-of-the-art black-box adversarial attacks. Section \ref{sec:preliminary} introduces the background knowledge. Section \ref{sec:method} explains our proposed framework and two adversarial attacks in detail. Section \ref{sec:experiments} shows the experimental results and their analysis. Section \ref{sec:discussion} discusses the limitations of our work. Finally, Section \ref{sec:conclusion} concludes the paper.

	\section{Related Work} \label{sec:relatedwork}
\subsection{Black-box Attacks}
In black-box attacks, only the input and output of the target model can be accessed and all internal information (e.g. parameters) is kept confidential. Moreover, the target model may only return the predictions of the top-$k$ categories rather than the complete probabilities for all categories \cite{ilyas18a}, which makes black-box attacks more challenging.

To address the problems, most existing methods regard all pixels as the same important and try to perturb all pixels of the input. Recently, a black-box attack \cite{li2019adversarial} only perturbs the foreground of the input because the background does not affect the prediction results. Observing this, we classify existing black-box adversarial perturbation into two categories: global perturbation and local perturbation.

\noindent\textbf{Global Perturbation.}
There are two main kinds of techniques for global perturbations in black-box attacks: gradient estimation and random search. Gradient estimation-based attacks \cite{chen2017zoo,ilyas2018black,bhagoji2018practical,du2018towards,tu2019autozoom,ilyas2018prior} follow the idea in white-box attacks to minimize an object function $J$ (e.g. mean square error) by changing the input $\rm{\textbf{x}}$. To guide the change, the gradient $\nabla_{\rm{\textbf{x}}} J$ is often needed. Since it cannot be calculated by backward propagation as in white-box attacks, gradient estimation-based attacks usually estimate $\nabla_{\rm{\textbf{x}}} J$ with a finite-difference method \cite{chen2017zoo} as follows:
\begin{equation}
\nabla_{\rm{\textbf{x}}} J=\begin{bmatrix}
\nabla_{{\rm{\textbf{x}}}_{00}} & \cdots  & \nabla_{{\rm{\textbf{x}}}_{0w}}\\ 
\vdots & \vdots  & \vdots \\ 
\nabla_{{\rm{\textbf{x}}}_{h0}}& \cdots  &\nabla_{{\rm{\textbf{x}}}_{hw}} 
\end{bmatrix},
\label{eq:FD}
\end{equation}
where
\begin{equation}
\nabla_{{\rm{\textbf{x}}}_{mn}}=
\frac{J({\rm{\textbf{x}}}+\epsilon \textbf{e}_{mn},y)-J({\rm{\textbf{x}}}-\epsilon \textbf{e}_{mn},y)}{2\epsilon} 
\label{eq:FDX}
\end{equation}
$\textbf{e}_{mn}$ is a matrix with the same size of $\rm{\textbf{x}}$. The element in the $m$-th row and $n$-th column is 1, other elements are 0. $\epsilon$ is a small real value. $y$ is the real label of $\rm{\textbf{x}}$. Then we can update $\rm{\textbf{x}}$ with $\nabla_{\rm{\textbf{x}}} J$. 

Random search-based attacks  \cite{brendel2018decision,narodytska2017simple,su2019one,alzantot2019genattack,guo2019simple,zhang2019attacking} is a kind of heuristic approaches that follow the mechanisms of ``feedback-fine-tuning" and ``survival of the fittest". In general, they first produce some random candidate perturbations by randomly perturbing some pixels of the input. Then they select some elite candidate perturbations according to the corresponding fitness that relies on the target model's feedback to produce better candidate perturbations. The whole process is repeated until there is a candidate that successfully fools the target model. The random search-based attacks are regarded as global attacks in this paper because all pixels are equally important when generating random perturbations.

One big drawback of global perturbations is their huge query overhead, especially the gradient estimation-based attacks, since they consider too many useless pixels (e.g. pixels of background). What's more, global perturbations modify the smoother background that leads the perturbation is more easily to be detected \cite{9157746}.

\noindent\textbf{Local Perturbation.}
Li \textit{et al.} \cite{li2019adversarial} proposed to only perturb the salient object in black-box attacks. Different from previous global perturbations, the foreground of the input is more significant for the prediction than the background. To determine the foreground, they assume there is a well-trained semantic segmentation model that can accurately determine the foreground and background. After identifying the foreground, they launch attacks based on random search by only modifying random pixels of the foreground with the same value.

Compared with randomly perturbing all pixels of clean examples, local perturbation attacks improve the query efficiency. However, Li \textit{et al.}'s work \cite{li2019adversarial} also has limitations. First, they use an expensive semantic segmentation model that needs higher quality datasets and more computational resources than the target model. Second, although they reduce the number of queries by local perturbation, the attack still needs huge queries to fool the targeted model. Third, the perturbations are easy to cause visible noises that would be detected by the defender.

\subsection{Query Reduction}
Query efficiency is a big challenge in black-box attacks. Since only the input and output of the target model can be accessed, one has to interact with the target model many times and acquire useful information from the feedback. To improve the query efficiency, some query reduction techniques are proposed. Here, we list four main kinds of existing state-of-the-art techniques: sharing queries, agent model, model ensemble, and hybrid. 

\noindent\textbf{Sharing Queries.}
As described above, applying Formula (\ref{eq:FDX}) to each pixel is very inefficient. To solve this problem, Bhagoji \textit{et al.} \cite{bhagoji2018practical} proposed to divide all pixels into small groups and each group contains a certain number of pixels. For each group, they use the directional derivative to replace the real derivative so that all pixels in one group share the same queries. As result, the new query times are reduced compared with the original attack. Based on sharing queries, Tu \textit{et al.} \cite{tu2019autozoom} further reduced query times with an auto-encoder.

%Following this idea, Tu \textit{et al.} \cite{tu2019autozoom} further proposed to query for a small size image first and then amplify perturbations with a pre-trained auto-encoder.

\noindent\textbf{Agent Model}
Papernot {et al.} \cite{papernot2017practical} proposed to train an agent model with soft labels \cite{hinton2015distilling} to approximate the target model's behaviors. This is designed to enhance the transferability of adversarial examples between the agent model and the target model. After training, adversarial examples can be produced through the agent model in a white-box way. Therefore, except querying the target model for checking whether the adversarial example succeeds, there is no need to interact with the target model anymore. However, training an agent model needs a huge number of queries to obtain these soft labels. Besides, the soft labels cannot be obtained when the target model does not return all probabilities for all categories.
%  But the training itself also needs many queries to acquire the soft labels that are the complete outputs of the target model. So just to attack a few clean examples, this scheme is not economical. In addition, the soft labels can not be obtained while we assume the output is limited. 

\noindent\textbf{Model Ensemble.}
Liu \textit{et al.} \cite{Yanpei2017delving} proposed to utilize multiple local models that are well-trained independently for the same task as the target model to enhance the transferability of adversarial examples. They claim that if an adversarial example cheats all the local models concurrently, there is a high probability for the adversarial example to fool the target model. With the increase of the local models, the success rate of the attack increases. Compared with training the agent model with soft labels, training the multiple local models in the model ensemble independently avoid interacting with the target model that saves the query overhead. However, it is hard to get such well-trained models and the model ensemble is compute-intensive, which is not applicable when computing resources are limited.

\noindent\textbf{Hybrid.}
Hybrid attack is first proposed in \cite{suya2020hybrid} that combines model ensemble and gradient estimation-based attacks. This approach also needs multiple local auxiliary models that are used to produce a universal adversarial example first. If the adversarial example fails, it will be further perturbed by gradient estimation-based attacks.  During estimating, all inputs and outputs (soft labels) are recorded to further train the $U$ local modals like that in \cite{papernot2017practical}. Due to the strong transferability of adversarial examples, the failed adversarial example can be closer to the target model's decision boundary than the corresponding clean example. Thus, fewer iterations are needed to optimize the objective function in gradient estimation-based attacks. However, this approach also faces the problems of both model ensemble and agent model techniques.

	\section{Preliminary} \label{sec:preliminary}

\subsection{Threat Model} 
We target black-box attacks in this paper and we can only access the target model through APIs. Besides, we also do not know any internal information about the target model. Further, we limit the output of the target model: the target model only returns the top-1 category and its probability, where the existing works \cite{papernot2017practical,suya2020hybrid} fail. These assumptions make our attacks more practical in real scenarios. 

We assume we can obtain a cheap or public available model for the same task as the reference model. Similar assumptions can be found in \cite{suya2020hybrid,Yanpei2017delving}. In Section \ref{sec:experiments}, we also explore the scenario in which we train the reference model by ourselves with a small dataset if there is no such an openly available model. It worth noting that the reference model is not necessary to have a competitive performance compared with the target model. As for the adversarial goal, we aim at non-targeted attacks with local perturbations, in which we only perturb the salient object to make the target model outputs an arbitrary wrong prediction. 

\subsection{Model Interpreters} \label{sec:preliminary:mi}
\begin{figure}[t]
	\centering	
	\includegraphics[width=0.8\columnwidth]{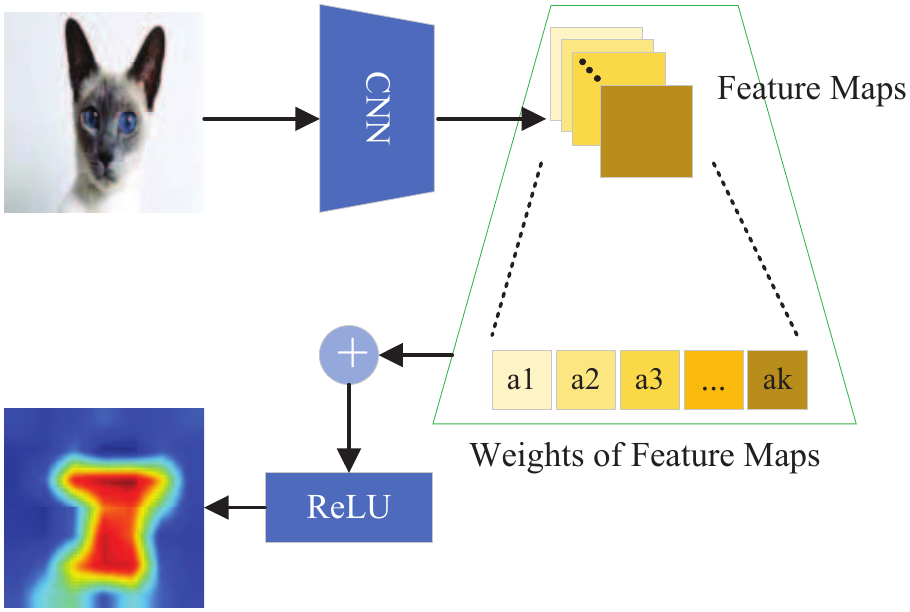}
	\caption{The pipeline of Grad-CAM.}
	\label{fig:Grad-CAM}
\end{figure} 
Model interpreters \cite{zhou2016learning,selvaraju2017grad,ribeiro2016should,8354201,omeiza2019smooth} are designed to identify which parts of the input the model's predictions rely on most. Without loss of generality, we choose Grad-CAM \cite{selvaraju2017grad} to determine the local important areas. It is an improved version of CAM \cite{zhou2016learning} and its biggest advantage, compared with CAM, is that it does not need to change the architecture of the model to be explained. 

For any model $F$ and any category $c$ of interest, Grad-CAM first calculates a weight $\alpha_{b}^{c}$ for each feature map $A^b$ of the last convolution layer of $F$. Then Grad-CAM calculates the weighted sum over all features maps with corresponding weights. Next, the ReLU function is applied to the weighted sum for filtering out negative effects. Finally, a salience map $SM$ is produced that reflects the importance of each pixel of the input to the target model. The whole process is illustrated in  Fig. \ref{fig:Grad-CAM} and can be formulated as 
\begin{equation}
	SM=ReLU\left(\sum_b \alpha^c_b A^b\right),
	\label{eq:grad}
\end{equation}
where 
\begin{equation}
  \alpha_{b}^{c}=\frac{1}{|A^b|}\sum_{m}\sum_{n}\frac{\partial {F}_c(\rm{\textbf{x}})}{\partial A^b_{mn}}.
\end{equation}
${F}_c(\rm{\textbf{x}})$ is the output of category $c$. $A^b_{mn}$ is the value of the $b$-th feature map $A^b$ at $(m,n)$. $|A^b|$ represents the total number of elements in $A^b$. 

\subsection{Image Fidelity Assessment}
$L_p$ norms are widely used as the image fidelity assessment (IFA) metric to measure the distance between original images and the corresponding adversarial examples. $p$ usually equals to 0, 1, 2, or $\infty$. However, Fezza \emph{et al}. \cite{fezza2019perceptual} claimed that the distance expressed by $L_p$ is quite different from the perception of human eyes. So we introduce other IFA metrics in this paper.

The first is the peak signal-to-noise ratio (PSNR). PSNR is defined as
\begin{equation}
	{\rm{PSNR}}=10*log_{10}\frac{GL^2}{{\rm{MSE}}},
	\label{eq:PSNR}
\end{equation}
where
\begin{equation}
	{\rm{MSE}}=\frac{1}{wh}\sum_{i=0}^{w-1}\sum_{j=0}^{h-1}{\left ( I_{ij} - J_{ij} \right )}^2.
\end{equation}
$I$ and $J$ represent two images of size $w\times h$ and $GL$ represents the gray levels.

Besides PSNR, we also consider the most apparent distortion (MAD) \cite{larson2010most}. For quantifying the similarity between clean examples and the corresponding adversarial examples, MAD is closest to the human eye perception among existing state-of-the-art metrics \cite{fezza2019perceptual}, including $L_p$ and $PSNR$. The calculation process of MAD can be summarized as five steps: 
\begin{enumerate}
	\item Compute the visible distortion map based on luminance images.
	\item Combine the visibility map with local errors by dividing the whole image into small blocks.
	\item Decompose both the distorted and original images into a set of subbands using a log-Gabor filter.
	\item Calculate different basic statistics (e.g. local standard deviation, local skewness, and local kurtosis) of each subband and combine to get the high order statistics.
	\item Calculate the adaptive blending score.
\end{enumerate}

	\section{Methodology} \label{sec:method}
In this section, we first illustrate the design strategies and the overview of our local and query-efficient black-box attack framework. Then, we present the details of generating adversarial examples with both gradient estimation and random search techniques.
\subsection{Framework Overview}
\begin{figure}[t]
	\centering
	\subfigure[VGG16]{\includegraphics[height=0.8in]{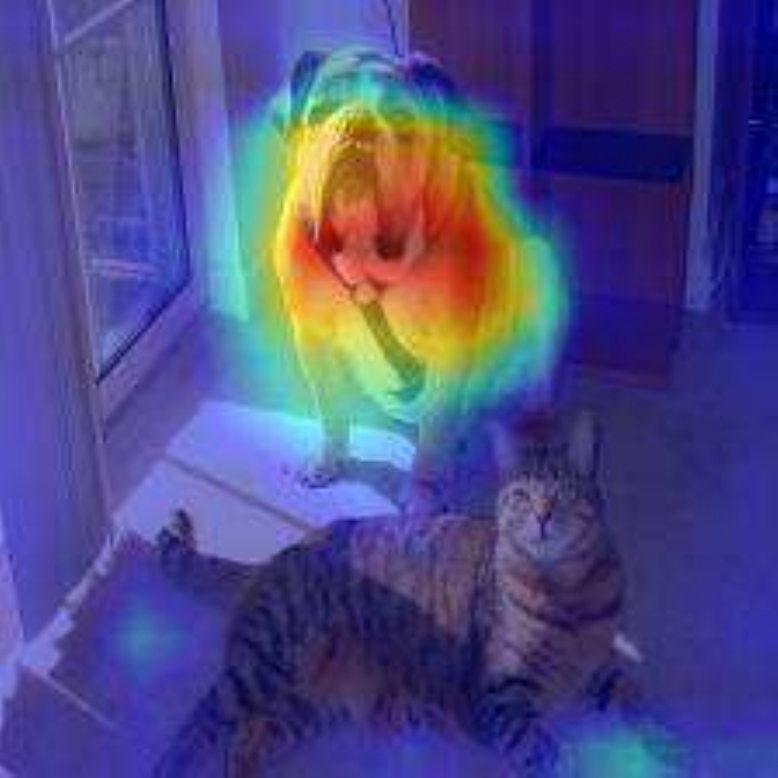}}
	\subfigure[VGG19]{\includegraphics[height=0.8in]{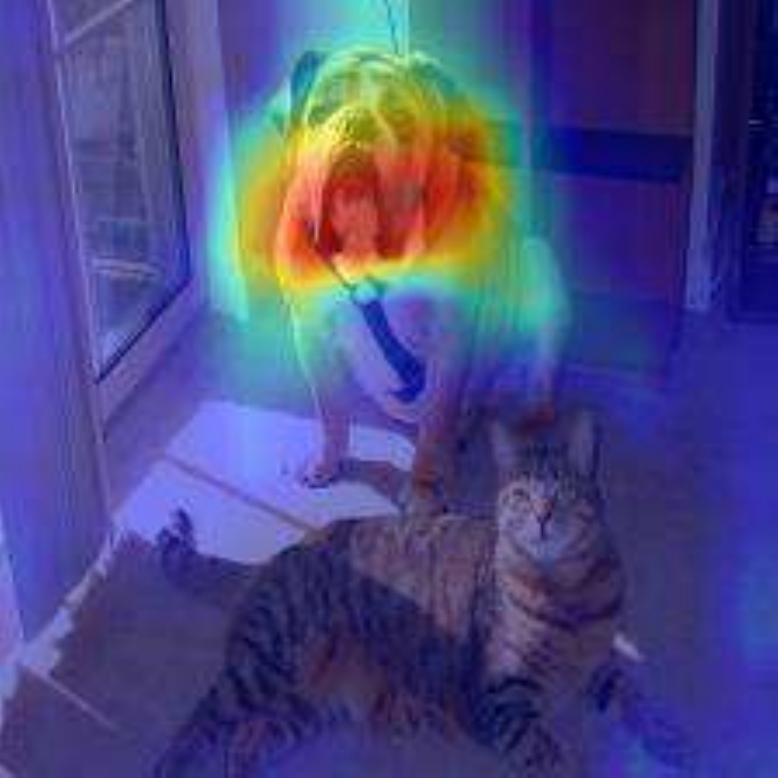}}
	\subfigure[ResNet50]{\includegraphics[height=0.8in]{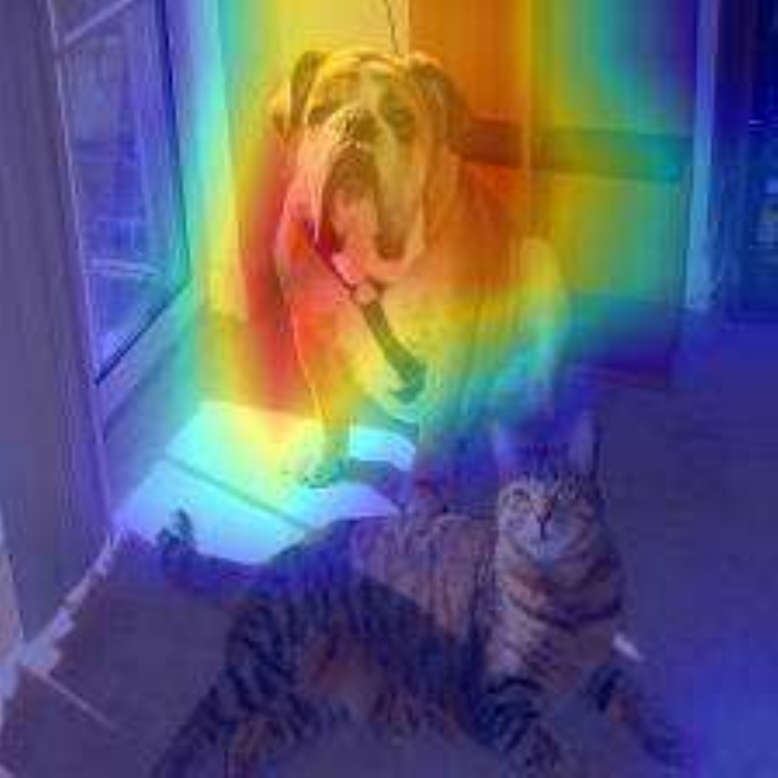}}	
	\subfigure[Inception-V3]{\includegraphics[height=0.8in]{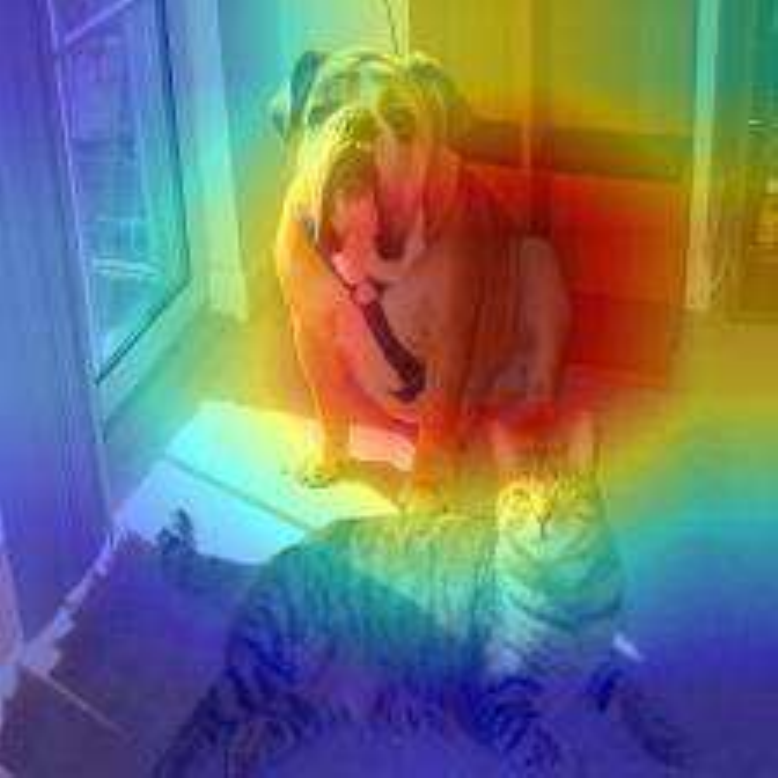}}
	\caption{Visualization of model interpretations using Grad-CAM \cite{selvaraju2017grad} on four well-known models: VGG16 \cite{simonyan2014very}, VGG19 \cite{simonyan2014very}, ResNet50 \cite{he2016deep}, and Inception-V3 \cite{szegedy2016rethinking}.}
	\label{fig:MI}
\end{figure}
\begin{figure*}[htbp]
	\centering
	\includegraphics[width=\textwidth]{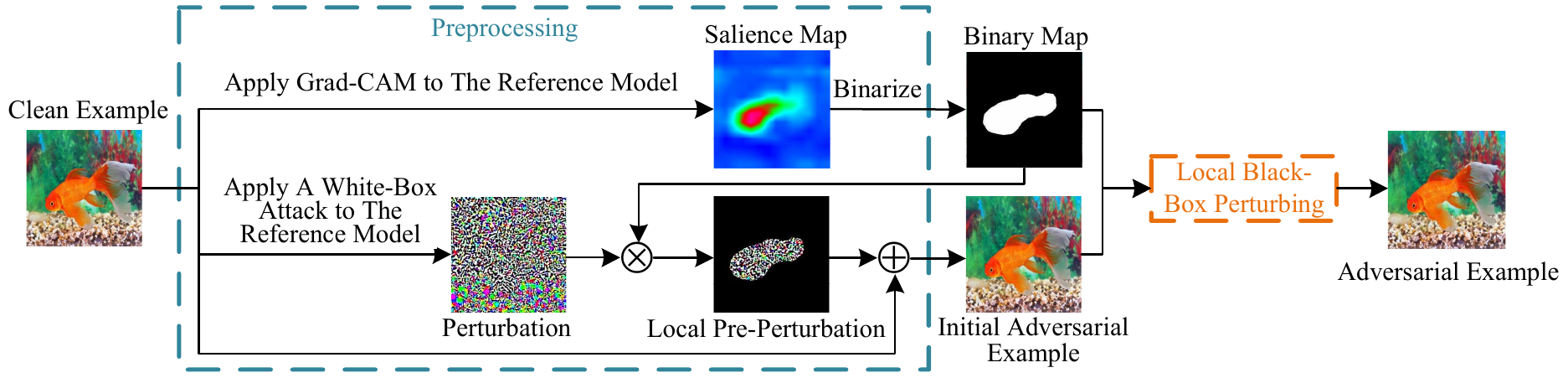}
	\caption{Pipeline of our local and query-efficient black-box attack framework. There are two phases: preprocessing and local black-box perturbing. The preprocessing phase consists of discriminative localization and pre-perturbation. We apply Grad-CAM to a reference model to determine the discriminative areas in a given clean example (e.g. the white area(s) of the binary map) due to the transferability of model interpretations. This is designed for forcing all subsequent perturbations to target only the salient object. Besides the binary map, the preprocessing also outputs an initial adversarial example with a local pre-perturbation. $\otimes$ and $\oplus$ represents element-wise multiplication and add operations. The initial adversarial example may successfully fool the target model due to the transferability of adversarial examples. Otherwise, the local black-box perturbing phase is triggered, in which black-box attacks are used to further perturb the initial adversarial example rather than the clean example because the initial adversarial example is a better start point.}
	\label{fig:mi-bba}
\end{figure*}

\noindent\textbf{Transferability.} The proposed framework is mainly based on our observations of the transferability of model interpretations and adversarial examples. For model interpretations, we find that there is a large overlap among the model interpretations even the corresponding models have different architectures and parameters (as shown in Fig. \ref{fig:MI}). The reason is that the visual objects for classification are always relatively fixed. Some works \cite{zhou2016learning,NIPS2017_0060ef47} have empirically shown that model interpreters can be used for weakly supervised object localization that only relies on category labels not bounding box information. This also indicates the transferability property of model interpretations. With the property, we can adopt a cheap or publicly available model as the reference model and accurately identify the discriminative areas of clean examples.

For adversarial examples, a lot of existing works have proven the transferability among different models \cite{szegedy2014,bhagoji2018practical,Yanpei2017delving}. Although the transferability of local perturbation of the discriminative areas is not guaranteed, a pre-perturbed example may be closer to the target model's decision boundary than the clean one. Thus, this reduces the number of queries for approaching the boundary and provides an advantage for the subsequent attack.

\noindent\textbf{Design Strateges.}
Most existing black-box attacks generate adversarial examples by global perturbations and perturb clean examples from scratch by interacting with the target model, which requires a huge number of queries to obtain satisfactory perturbations. In contrast, we adopt the following two transferability-based design strategies to produce local perturbations and to significantly improve the query efficiency. First, we take advantage of \emph{the transferability of model interpretations} to accurately identify the discriminative areas of clean examples. Since the prediction is heavily affected by the discriminative pixels, perturbing the discriminative pixels can be more useful than the other pixels. Second, inspired by \emph{the transferability of adversarial examples}, we further reduce the complexity of the interaction between the adversary and the targeted model by pre-perturbing the discriminative pixels. This ensures that the pre-perturbed samples are closer to the decision boundary of the target model, thus decreasing the queries needed. All the two strategies can be achieved through the same reference model.

\noindent\textbf{Pipeline.} Fig. \ref{fig:mi-bba} illustrates our attack framework which consists of two phases: preprocessing and local attack. In the preprocessing phase, given a clean example, we identify the discriminative area(s) that determines the most important part of the clean example to the target model. Here, we adopt Grad-CAM \cite{selvaraju2017grad} to the reference model and obtain an accurate salience map. Due to the transferability of model interpretation, the map is also suitable for the target model. Then, we binarize the salience map into a binary map to get the discriminative areas (e.g the white area of the binary map). Second, we take the advantage of the transferability of adversarial examples and use the reference model to produce an initial adversarial example by pre-perturbing the clean example locally with an arbitrary white-box adversarial attack. If the initial adversarial example can fool the target model, we successfully achieve our goal. Otherwise, we employ the local black-box perturbing phase to further perturb the initial adversarial example in a black-box way.

\subsection{Preprocessing}
The preprocessing is designed for identifying the attack areas for local perturbations and improving the final query efficiency in the local black-box perturbing phase. According to the pipeline in Fig. \ref{fig:mi-bba}, we make full use of one reference model and the two kinds of transferability to realize two goals concurrently: discriminative localization and pre-perturbation. It worth noting that the use of auxiliary models in previous work \cite{li2019adversarial,papernot2017practical,Yanpei2017delving,suya2020hybrid} is limited to one of the two goals.

Before the attack, we only need one arbitrary model for the same task as the target model. It can be a publicly available model or be trained by ourselves. Compared with the usage of multiple different models in \cite{Yanpei2017delving,suya2020hybrid}, we reduce the number of auxiliary models to one so that the requirement is simplified and it is more likely to execute attacks with limited computational resources. Compared with the semantic segmentation model used in \cite{li2019adversarial}, our reference model is easier to be trained even when no public model is available because training our reference model only needs category-level labels while training the semantic segmentation needs pixel-level labels. In addition, the semantic segmentation model can only provide discriminative localization.

Given a clean example, we first locate the discriminative areas of the clean example ${\rm {\textbf{x}}}$ based on the transferability of model interpretations. In experiments, we apply Grad-CAM \cite{selvaraju2017grad} to the reference model to get the salience map $SM$ according to Formula (\ref{eq:grad}), where $F$ and $c$ are the reference model $f_r$ and the real category $y$ of the clean example. We resize $SM$ to the same size of $\rm{\textbf{x}}$ and afterward we binarize it to a binary map $BM$ with the average of the resized $SM$ as the threshold. Specifically, if $SM_{mn}$ is greater than or equal to the average, $BM_{mn}$ is 1. Otherwise, $BM_{mn}$ is 0, where $(m, n)$ is the coordinate of a pixel.

After identifying discriminative areas, instead of directly querying the targeted model, we propose to produce a better start point based on the transferability of adversarial examples to reduce the number of queries. In experiments, we refer to BIM \cite{kurakin2016adversarial} to produce the initial adversarial example, which is a simple and effective white-box attack. Unlike BIM perturbs all pixels, we only change the discriminative pixels. We denote the new perturb way as Local-BIM that can be described below:
\begin{equation}
	{\rm{\textbf{x}}}_{t+1}=clip\left({\rm{\textbf{x}}}_{t}+\epsilon_1*(sign(\nabla_{{\rm{\textbf{x}}}_{t}}J({\rm{\textbf{x}}}_{t},y;\theta_{f_r}))\otimes BM )\right ) \label{eq:bim}
\end{equation}
where ${\rm{\textbf{x}}}_{t}$ is the input in the $t$-th iteration. $\epsilon_1$ controls the amount of the perturbation of each pixel. $sign$() is
\begin{equation}
	sign(x)=\left\{\begin{matrix}
		1,& {\rm{if}} \;\; x>0\\
		0,& {\rm{if}} \;\; x=0\\
		-1,& {\rm{if}} \;\; x<0
	\end{matrix}\right.
	\label{eq:sign}
\end{equation}
$\theta_{f_r}$ is the parameters of $f_r$. $\nabla_{{\rm{\textbf{x}}}_{t}}J({\rm{\textbf{x}}}_{t},y;\theta_{f_r})$ represents the partial derivatives of $J({\rm{\textbf{x}}}_{t},y;\theta_{f_r})$ with respect to ${\rm{\textbf{x}}}_{t}$. One can easily calculate $\nabla_{{\rm{\textbf{x}}}_{t}}J({\rm{\textbf{x}}}_{t},y;\theta_{f_r})$ through back propagation. $\otimes$ is the element-wise multiplication that ensures perturbations only occur in the discriminative areas. $clip$() is used to ensure each element of ${\rm{\textbf{x}}}_{t+1}$ in a legal range. We iterate the whole process at most $T_1$ times. After each iteration, we query the target model with the updated example to verify whether it cheats the target model. If it succeeds, the whole attack is ended. Otherwise, after $T_1$ iterations, the second phase is triggered with $BM$ and the latest perturbed example ${\rm{\tilde {\textbf{x}}}}_{ini}$ (the initial adversarial example). The whole preprocessing is summarized in Algorithm \ref{al:pre}.

\begin{algorithm}[t!]
 	\caption{\textbf{Preprocessing}}
 	\label{al:pre}
 	\SetAlgoLined
 	\SetKwInOut{Input}{Input}
	\SetKwInOut{Output}{Output}
 	\Input{Clean example $\rm\textbf{x}$, real label $y$,  target model $f$, reference model $f_r$, iteration parameter $T_1$, perturbation parameter $\epsilon_1$.}
 	\Output{Initial adversarial example ${\rm{\tilde {\textbf{x}}}}_{ini}$, binary map $BM$.}

	$SM\leftarrow Grad\_CAM({\rm\textbf{x}},y,f_r)$\;
	$SM\leftarrow Resize(SM,({\rm\textbf{x}}.width,{\rm\textbf{x}}.height))$\;
	$BM\leftarrow Binarize(SM)$\;
	${\rm{\tilde {\textbf{x}}}}_{ini}\leftarrow{\rm\textbf{x}}$\;
	\For{$iter\leftarrow 1$ to $T_1$}{
		${\rm{\tilde {\textbf{x}}}}_{ini}\leftarrow Local\_BIM({\rm{\tilde {\textbf{x}}}}_{ini},y,f_r,BM)$;
		$isSucceed\leftarrow$Check whether ${\rm{\tilde {\textbf{x}}}}_{ini}$ succeeds cheat $f$\;
		\If{$isSucceed$}{
				End attack with ${\rm{\tilde {\textbf{x}}}}_{ini}$ as the final adversarial example\;
		}
	}
 	\Return ${\rm{\tilde {\textbf{x}}}}_{ini}$, $BM$\;
\end{algorithm}

\subsection{Local Black-Box Perturbing}\label{sec:method:c}
In this phase, with the transferability of adversary examples, we locally perturb ${\rm{\tilde {\textbf{x}}}}_{ini}$ rather than ${\rm{\textbf{x}}}$ in a black-box way because ${\rm{\tilde {\textbf{x}}}}_{ini}$ can save queries for approaching the decision boundary of $f$. The general process of the local black-box perturbing phase can be found in Algorithm \ref{al:lbba}, where the function $Local\_Perturb()$ represents a specific local black-box perturbing process that can be one of gradient estimation-based or random search-based adversarial attacks. In the following, we introduce two local black-box attacks based on gradient estimation and random search, respectively, to produce the final adversarial examples.

\begin{algorithm}[t!]
	\caption{\textbf{Local\_Black\_Box\_Attack}}
	\label{al:lbba}
	\SetAlgoLined
	\SetKwInOut{Input}{Input}
	\SetKwInOut{Output}{Output}
	\Input{Initial adversarial example ${\rm{\tilde {\textbf{x}}}}_{ini}$, binary map $BM$, real label $y$, target model $f$, other hyper-parameter set $\lambda$.}
	\Output{Final adversarial example ${\rm{\tilde {\textbf{x}}}}$.}	
	${\rm{\tilde {\textbf{x}}}}\leftarrow Local\_Perturb({\rm{\tilde {\textbf{x}}}}_{ini},y,f,BM,\lambda)$\;
	$isSucceed\leftarrow$Check whether ${\rm{\tilde {\textbf{x}}}}$ succeeds cheat $f$\;
	\If{$isSucceed$}{
		\Return ${\rm{\tilde {\textbf{x}}}}$\;
	}
	\Else{
		\Return None
	}
\end{algorithm}

\subsubsection{\textbf{Local Black-Box Attack with Gradient Estimation}}\label{sec:method:c:a}
Our preprocessing phase can be directly applied to existing gradient estimation-based schemes \cite{chen2017zoo,bhagoji2018practical} and improve the query efficiency. Considering the target model only returns the probability of the top-1 category and our attack framework is of local perturbation, we revise these gradient estimation-based schemes that need extra outputs to estimate gradients and perturb globally. Here, we take \cite{bhagoji2018practical} as an example and revise the attack for our framework.

In \cite{bhagoji2018practical}, they approximate BIM with finite difference. Here, we refer to the way to approximate Local-BIM. In particular, we estimate $\nabla_{{\rm\tilde{{\textbf{x}}}}_{{ini}_{mn}}}$ through
\begin{equation}
	 \nabla_{{\rm\tilde{{\textbf{x}}}}_{{ini}_{mn}}}=\left\{\begin{matrix}
		\frac{f_y({\rm{\tilde{\textbf{x}}}}_{ini}+\epsilon e_{mn})-f_y({\rm{\tilde{\textbf{x}}}}_{ini}-\epsilon e_{mn})}{2\epsilon} & BM_{mn}=1\\
		0& otherwise
	\end{matrix}\right.
	\label{eq:FDXmn}
\end{equation}
According to our philosophy, we do not estimate for all pixels but only the discriminative pixels. We set $\nabla_{{\rm\tilde{{\textbf{x}}}}_{{ini}_{mn}}}=0$ when $BM_{mn}=0$ to ensure that we do not have to waste queries for unimportant pixels and the final perturbation only occurs in the discriminative areas. In addition, all calculation only relies on the output of the real category $y$. If the top-1 category is not $y$, it means we have found an adversarial example. Otherwise, the real category $y$ is the top-1 category and $f_y$ is the corresponding probability. It is consistent with the hypothesis that the target model only returns the result of the top-1 category. After we get the whole estimation according to Formula (\ref{eq:FDXmn}) that reflects the growth direction of $f_y$, we update ${\rm\tilde{{\textbf{x}}}}_{{ini}}$ along the opposite direction to reduce the probability. We repeat this process at most $T_2$ times:
\begin{equation}
	{\rm\tilde{{\textbf{x}}}}_{{ini}_{t+1}}=clip\left({\rm\tilde{{\textbf{x}}}}_{{ini}_{t}}+\epsilon_2*sign\left (-\nabla_{{\rm\tilde{{\textbf{x}}}}_{{ini}_{t}}} f_y \right)\right)
	\label{eq:BIM}
\end{equation}
${\rm\tilde{{\textbf{x}}}}_{{ini}_{0}}={\rm\tilde{{\textbf{x}}}}_{{ini}}$. Similar to \cite{bhagoji2018practical}, one can also use random grouping or PCA to further improve query efficiency by sharing queries.

\subsubsection{\textbf{Local Black-Box Attack with Random Search}}
Several random search-based black-box attacks \cite{brendel2018decision,narodytska2017simple,li2019adversarial,su2019one,alzantot2019genattack,guo2019simple,zhang2019attacking} have been proposed and most of them are of global perturbation. Although  Li et al. \cite{li2019adversarial} determine the foreground for locally perturbing, their attack has some limitations. First, Li et al. \cite{li2019adversarial} start the black-box perturbing from clean examples and thus require a huge amount of queries. Second, they set all the perturbed pixels with the same value, which could affect the quality of the final adversarial examples. Next, we revise their method to realize the local black-box perturbing phase and overcome the limitations.

In particular, we start with the initial adversarial example rather than the clean example for higher query efficiency. Moreover, we produce the $K$ candidate adversarial examples by adding a perturbation $\epsilon_3$ to different discriminative pixels of $K$ copies of ${\rm\tilde{{\textbf{x}}}}_{{ini}}$ for better quality rather than set them with the same value. After the $clip$ operation, all copies are candidate adversarial examples, and are denoted as ${\rm{\tilde {\textbf{x}}}}^i_{can}$ ($i=1, 2, ..., K$). Then, we test the importance of each modification in the candidates by feeding the candidates into the target model. According to the feedback $f_y({\rm{\tilde {\textbf{x}}}}^i_{can})$, we sort the $K$ candidates in the ascending order. Next, we combine all perturbations of the top $R$ elite candidates ($R < K$) to update the initial adversarial example, and it becomes a new candidate. We repeat the whole process until any candidate fools the target model or the maximum number of iterations $T_3$ is reached.
    \section{Experiments} \label{sec:experiments}

\subsection{Experimental Setup}
\noindent\textbf{Dataset}. All the target models are trained on ImageNet. In addition, to simulate attacks in real scenes, we randomly collect 410 images through Google for attacking. These images consist of animals, transports, and traffic signs.
    
\noindent\textbf{Model}. Our approach is applicable to various models. We choose two popular models: Inception-V3 \cite{szegedy2016rethinking} and ResNet50 \cite{he2016deep} as the target classifiers. We use VGG16 \cite{karen2015very} as the reference classifier. The input size of ResNet50 and VGG16 is (224,224,3) and each channel of an image needs to be subtracted a fixed mean value before the image is fed into ResNet50 and VGG16. Inception-V3's input size is (299,299,3) and its input need to be normalized into the range of [-1,1] from [0,255] before the prediction. 

\noindent\textbf{Baselines and the Proposed Attacks}. 
We first introduce the gradient estimation-based attacks:
\begin{itemize}
	\item \textbf{GGE}: global gradient estimation (GGE) is a global attack that applies finite-difference to estimate gradient for all pixels. We choose IFD-xent \cite{bhagoji2018practical} as the GGE method and implement it with random grouping for query reduction.

	\item \textbf{MI-GE}: model interpretation-based gradient estimation (MI-GE) is a local attack that limits the perturbing areas of GGE by the transferability of model interpretations. Compared with GGE, MI-GE only estimates the gradient for discriminative pixels of the clean example. 
	
	\item \textbf{IAE\&MI-GE}: the proposed attack under our framework with initial adversarial examples and model interpretation-based gradient estimation (IAE\&MI-GE). It not only identifies discriminative areas but also produces an initial adversarial example before the estimating. IAE\&MI-GE also uses random grouping during the estimation process. 

\end{itemize}

Similarly, we list the random search-based baselines and the proposed attack:
\begin{itemize}
	\item \textbf{GRS}: global random search (GRS) is also a global attack. We refer to subject-based local search (SBLS) \cite{li2019adversarial} and remove its preprocessing phases to implement GRS. Thus, GRS randomly perturbs pixels of the whole image instead of only the foreground. Except for this point, the rest of GRS are consistent with SBLS.
	
	\item \textbf{SBLS}: SBLS \cite{li2019adversarial} is a local attack. It uses a semantic segmentation model to determine the foreground first. Then it starts perturbing with clean examples and produces candidate adversarial examples by randomly perturbing the pixels within the foreground. All perturbed pixels are set with the same value.
	\item \textbf{MI-RS}: we construct another local attack: model interpretation-based random search (MI-RS). Similar to MI-GE, MI-RS first determines discriminative areas by the transferability of model interpretations. To produce candidate adversarial examples, MI-RS randomly adds $\epsilon_3$ to different discriminative pixels. 
	\item \textbf{IAE\&MI-RS}: the proposed attack under our framework with initial adversarial example and model interpretation-based random search (IAE\&MI-RS). IAE\&MI-RS starts the random search process with an initial adversarial example and a binary map.
\end{itemize}

\subsection{Performance Metrics}\label{sec:experiments:criterion}
We employ three criteria for the evaluation of black-box attacks: number of quires (NoQ), success rate (SR), and quality of adversarial examples (QoAE). NoQ quantifies the number of querying on the target model in one whole attack, and a smaller value of NoQ indicates a better query efficiency. For random search-based attacks we test in experiments, NoQ also reflects the number of noises. Smaller NoQ means fewer noises. SR measures the rate of the number of successfully cheating the target model to the total number of attacks. Conventionally, SR is defined as 
\begin{equation}
\text{SR}=\frac{1}{N} \sum_{j}^{N}\mathbbm{1}\left(\mathop{\arg\max}\limits_{c}f_c({\rm{\tilde {\textbf{x}}}}^j) \neq y^j\right)*100\%
\label{eq:sr}
\end{equation}
where $N$ represents the total number of clean examples and it is equal to 410 in our experiment. ${\rm{\tilde {\textbf{x}}}}^j$ is the $j$-th adversarial example and $y^j$ is its real label. $\mathbbm{1}()$ is 1 while its input is true. Otherwise, it equals to 0. QoAE measures the similarity between clean examples and the corresponding adversarial examples. In experiments, we use MAD \cite{larson2010most} and PSNR to quantify the QoAE. PSNR is a well-known criterion and has been widely used in related studies. A larger PSNR value means better similarity. Meanwhile, we also use MAD to measure the QoAE, which is proved to be more consistent with the perception of the human visual system \cite{fezza2019perceptual}. Different from PSNR, a smaller MAD value represents better similarity. We report the average values of all results of NoQ and QoAE.

\subsection{Overall Results}
\begin{table*}[htbp]
	\caption{Overall evaluation of our proposed attacks and the baselines on the whole dataset when the target model is ResNet50}
	\label{tb:overall_resnet50}
	\centering
	\begin{tabular}{|c|c|c|c|c|c|c|c|c|}
		\hline
		\multicolumn{2}{|c|}{Category}  & \multicolumn{3}{c|}{Gradient Estimation}           & \multicolumn{4}{c|}{Random Search}                        \\ \hline
		\multicolumn{2}{|c|}{Method}    & IAE\&MI-GE & MI-GE & GGE  & IAE\&MI-RS    & MI-RS & SBLS & GRS \\ \hline
		\multicolumn{2}{|c|}{NoQ}       & \textbf{8377}       & 19124 & 35770                         & \textbf{223}           & 405      & 376      & 452                 \\ \hline
		\multicolumn{2}{|c|}{SR}    & 99.51\%      & 98.04\% & \textbf{100\%}                           & \textbf{97.80\%}         & \textbf{97.80\%}     &\textbf{ 97.80\%}     & \textbf{97.80\%}              \\ \hline
		
		\multirow{2}{*}{QoAE} & MAD     & 11.10      & 11.78 & \textbf{8.72}                          & \textbf{30.29}         & 44.91    & 37.04    &  51.65              \\ \cline{2-9}
		& PSNR    & 40.09      & \textbf{42.04} & 38.99                         & \textbf{36.44}         & 32.94    & 32.92    &32.49                \\ \hline
	\end{tabular}
\end{table*}

\begin{table*}[htbp]
	\caption{Overall evaluation of our proposed attacks and the baselines on the whole dataset when the target model is InceptionV3}
	\centering
	\label{tb:overall_inceptionv3}
	\begin{tabular}{|c|c|c|c|c|c|c|c|c|}
		\hline
		\multicolumn{2}{|c|}{Category}  & \multicolumn{3}{c|}{Gradient Estimation} & \multicolumn{4}{c|}{Random Search}       \\ \hline
		\multicolumn{2}{|c|}{Method}    & IAE\&MI-GE    & MI-GE   & GGE   & IAE\&MI-RS & MI-RS & SBLS  & GRS    \\ \hline
		\multicolumn{2}{|c|}{NoQ}       & \textbf{12627}         & 18186   & 65189          & \textbf{280}           & 716      & 678   & 895   \\ \hline
		\multicolumn{2}{|c|}{SR}     & \textbf{99.76\%}             & 99.02\%   & 99.27\%          & \textbf{92.68\%}         & 86.83\%    & 89.02\% & 83.66\% \\ \hline
		\multirow{2}{*}{QoAE} & MAD     &29.18               & \textbf{6.68}    & 9.27           & \textbf{48.45}         & 55.20    & 48.64 & 64.84 \\ \cline{2-9}
		& PSNR    & 34.33            & \textbf{42.65}   & 42.11          & 31.92         & \textbf{32.81}    & 32.75 & 31.94 \\ \hline
	\end{tabular}
\end{table*}
We first detail the key parameter settings used in this subsection. We set the $\epsilon_1$ and $T_1$ used for generating the initial adversarial examples as 1.5. and 5, respectively. When we use GGE, MI-GE and IAE\&MI-GE to launch attacks, $T_2$ is 10 and the random grouping size is 20. $\epsilon_2$ relies on the target model because the input range is different. For ResNet50 and Inception-V3, $\epsilon_2$ equals to 2 and 0.2, respectively. When we use GRS, SBLS, MI-RS and IAE\&MI-RS to attack the target model, $K=50$, $R=30$ and $T3=60$. Without losing generality, both SBLS and GRS modify all selected pixel into white (255,255,255). As for $\epsilon_3$ in MI-RS and IAE\&MI-RS, it also relies on the target model. We first use large $\epsilon_3$ ($\epsilon_3=255$ for ResNet50 and $\epsilon_3=2$ for Inception-V3) so that the changed pixels after $clip()$ are also white. Latter, we reduce $\epsilon_3$ to show better visual effect. Under the above experiment settings, we illustrate the experimental results in Table \ref{tb:overall_resnet50} and \ref{tb:overall_inceptionv3}.

\begin{figure}[t]
	\centering
	\subfigure[Clean]{\includegraphics[width=0.8in]{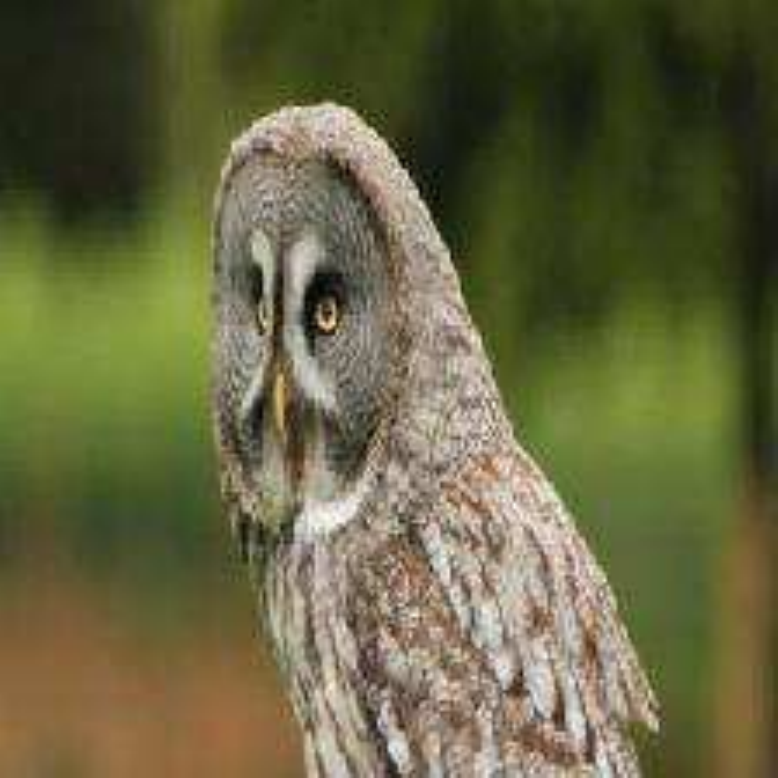}\label{fig:examples-of-different-methods:a}}
	\subfigure[IAE\&MI-GE]{\includegraphics[width=0.8in]{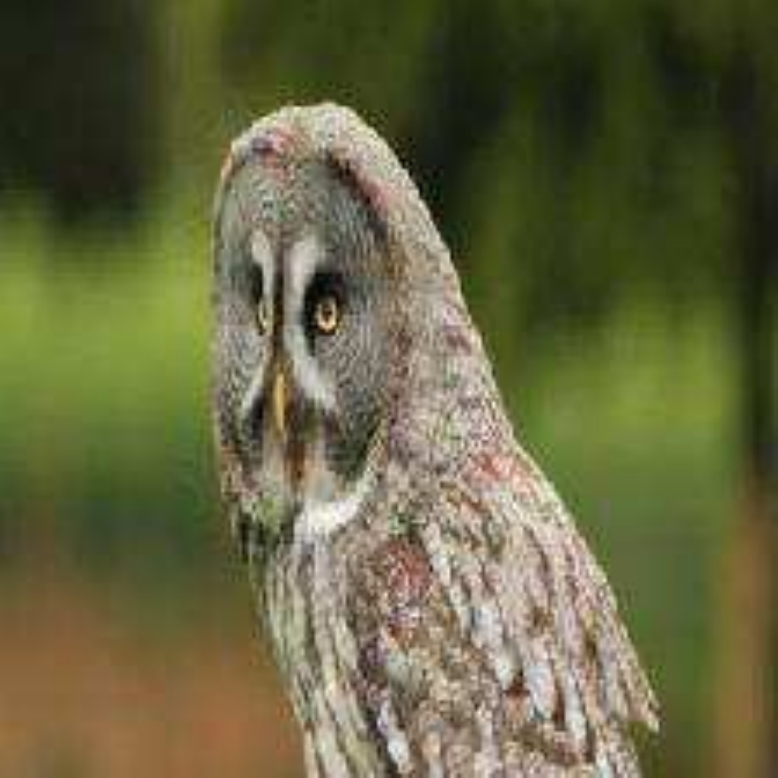}\label{fig:examples-of-different-methods:b}}	
	\subfigure[MI-GE]{\includegraphics[width=0.8in]{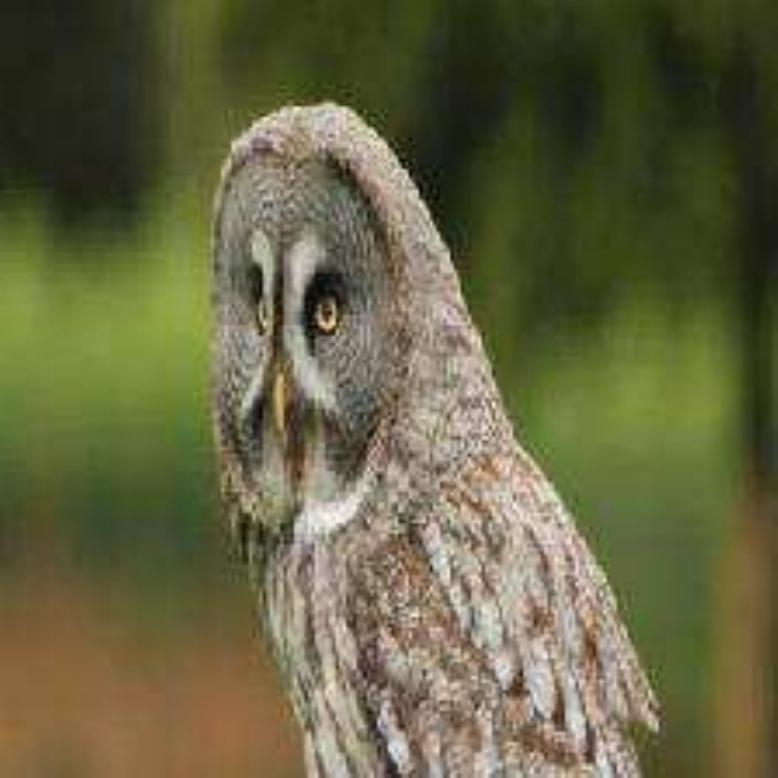}\label{fig:examples-of-different-methods:c}}
	\subfigure[GGE]{\includegraphics[width=0.8in]{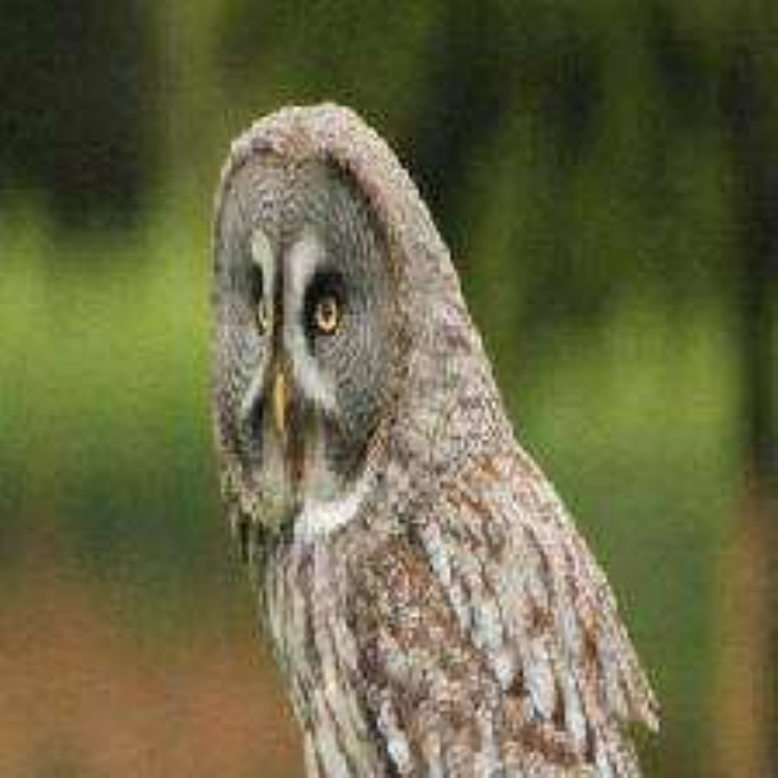}\label{fig:examples-of-different-methods:d}}
	\subfigure[IAE\&MI-RS]{\includegraphics[width=0.8in]{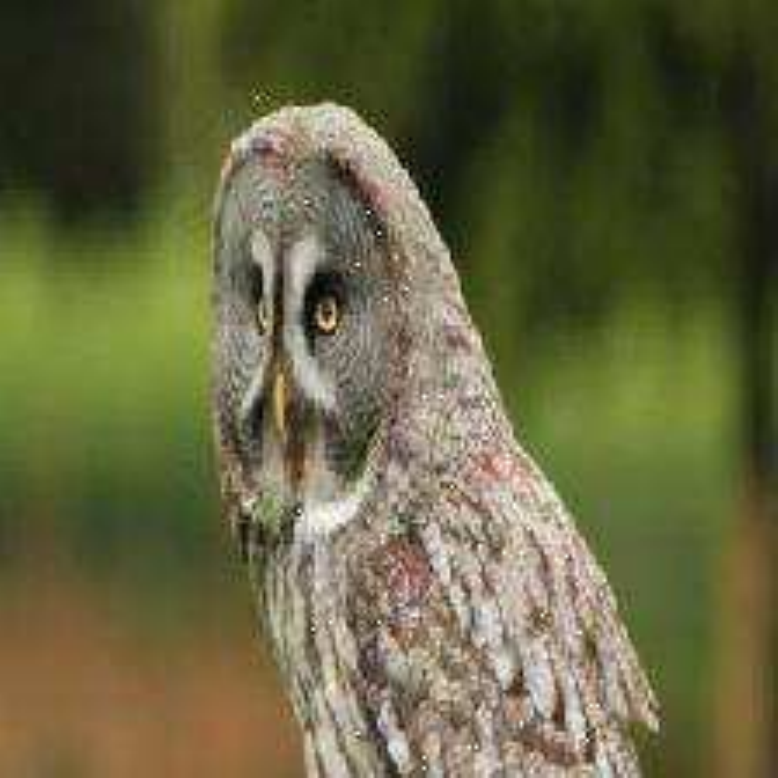}\label{fig:examples-of-different-methods:e}}
	\subfigure[MI-RS]{\includegraphics[width=0.8in]{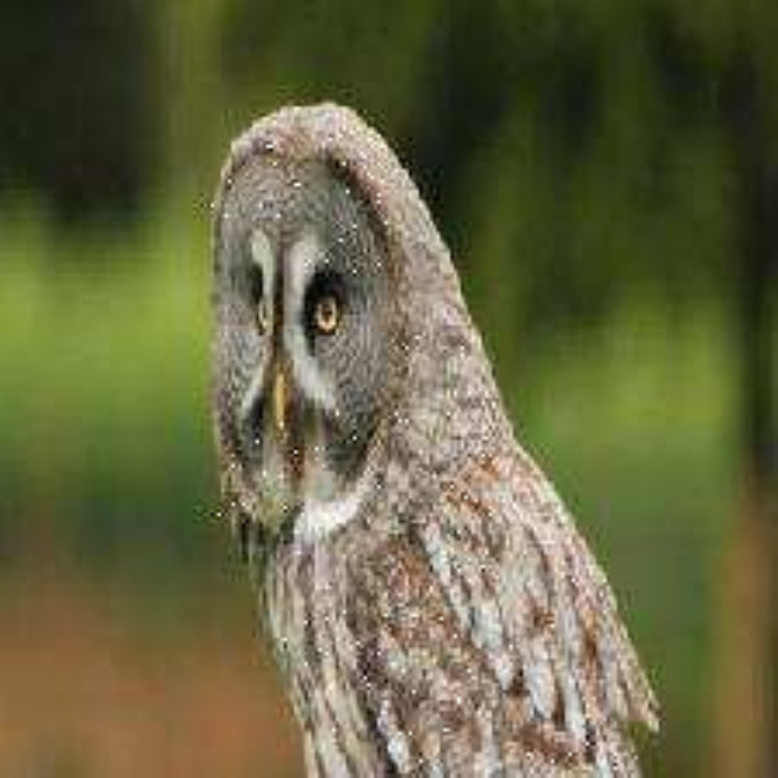}\label{fig:examples-of-different-methods:f}}
	\subfigure[SBLS]{\includegraphics[width=0.8in]{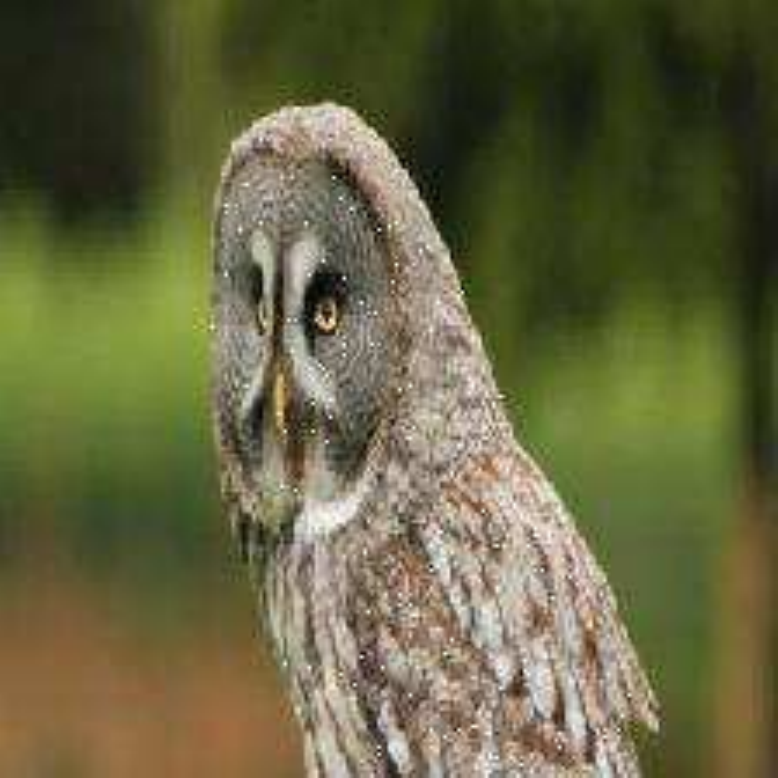}\label{fig:examples-of-different-methods:g}}
	\subfigure[GRS]{\includegraphics[width=0.8in]{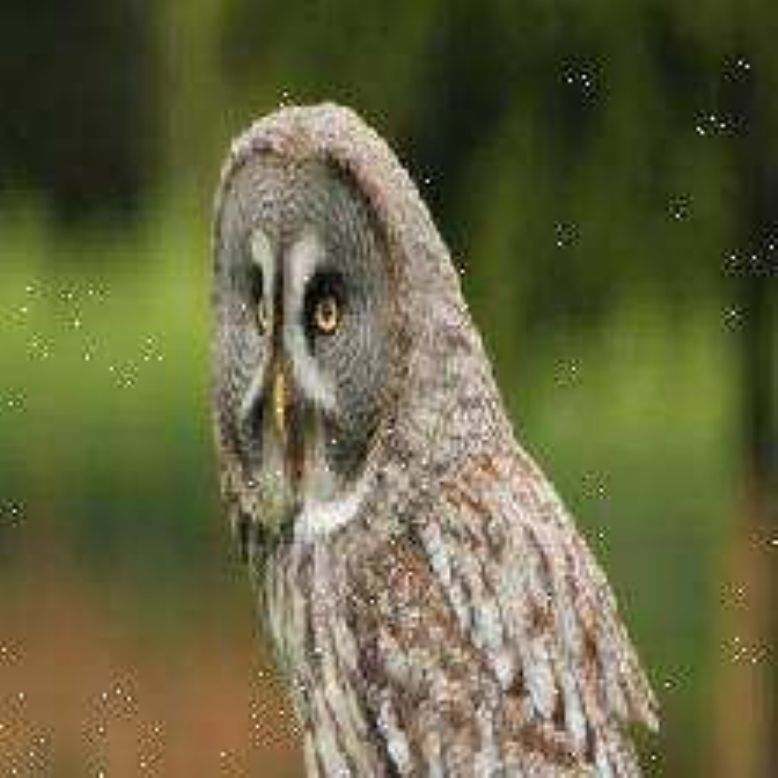}\label{fig:examples-of-different-methods:h}}
	
	\caption{Adversarial examples generated by different methods when the target model is ResNet50.  (a) Original image; (b) Adversarial image generated by IAE\&MI-GE (NoQ=4440, MAD=4.31, PSNR=38.93); (c) Adversarial image generated by MI-GE (NoQ=17745, MAD=2.22, PSNR=41.28); (d) Adversarial image generated by GGE (NoQ=45170, MAD=12.49, PSNR=37.31); (e) Adversarial image generated by IAE\&MI-RS (NoQ=94, MAD=13.89, PSNR=36.49); (f) Adversarial image generated by MI-RS (NoQ=357, MAD=25.63, PSNR=32.34); (g) is generated by SBLS (NoQ=306,MAD=21.99, PSNR=33.05); (h) Adversarial image generated by GRS (NoQ=514, MAD=83.22, PSNR=30.80).}
	\label{fig:examples-of-different-methods}
\end{figure}

We observe that our IAE\&MI-GE and IAE\&MI-RS achieve the best query efficiency in the corresponding type of attacks. Taking Table \ref{tb:overall_resnet50} as an example, compared with global ways (e.g. GGE and GRS), IAE\&MI-GE and IAE\&MI-RS reduce the NoQ by 76.58\% and 50.66\%, respectively. Meanwhile, we observe that only focusing on perturbing discriminative pixels (e.g. MI-GE, MI-RS, and SBLS) also improves the efficiency compared with global ways. Specifically, compared with GGE, MI-GE reduces the NoQ by 46.53\%. And compared with GRS, MI-RS and SBLS reduce the NoQ by 10.4\% and 16.81\%, respectively. However, the improvements in query efficiency brought by MI-GE, MI-RS, and SBLS are limited compared with those brought by IAE\&MI-GE and IAE\&MI-RS. In Fig. \ref{fig:examples-of-different-methods}, MI-GE and MI-RS cost 17745 and 357 queries respectively to produce the adversarial examples (Fig. \ref{fig:examples-of-different-methods:c} and \ref{fig:examples-of-different-methods:f}). After pre-perturbing, IAE\&MI-GE and IAE\&MI-RS only need 4440 and 94 queries (Fig. \ref{fig:examples-of-different-methods:b} and \ref{fig:examples-of-different-methods:e}).

It worth noting that SBLS that uses semantic segmentation models to determine foreground costs 29 fewer queries than MI-RS on average. However, this gap of NoQ is not obvious, and it also proves that it is feasible to use the transferability of model interpretations to identify discriminative areas.  

Besides query efficiency, we also care about SR and QoAE. If the SR value is low, the corresponding attack method may be non-attractive. When the target model is ResNet50, all tested methods achieve high SR. And when the target model is Inception-V3, the SR values of all gradient-estimation based attacks are higher than 99\%. However, among random search-based attacks, only IAE\&MI-RS exceeds 90\% SR. All results show that, compared with GGE and GRS, only focusing on the discriminative areas slightly reduces the SR for gradient estimation-based attacks but increases the SR for random search-based attacks. This is because GGE perturbs all pixels including discriminative pixels, but GRS may only perturb a few discriminative pixels and more useless pixels. Besides, pre-perturbing helps IAE\&MI-GE and IAE\&MI-RS increase the SR compared with MI-GE and MI-RS.

As for QoAE, in general, MI-GE, MI-RS, and SBLS respectively get better scores compared GGE and GRS because a lot of unnecessary perturbations on useless pixels are avoided. In Fig. \ref{fig:examples-of-different-methods:d}, the whole background becomes more blurred than that of Fig. \ref{fig:examples-of-different-methods:c}. And the white noises in the background in Fig \ref{fig:examples-of-different-methods:h} are easier to be perceived than those in the discriminative areas in Fig \ref{fig:examples-of-different-methods:f} and \ref{fig:examples-of-different-methods:g}. But the white noises in Fig \ref{fig:examples-of-different-methods:f} and \ref{fig:examples-of-different-methods:g} also influence the visual effect. So we reduce the $\epsilon_3$ for better quality in the next subsection. Meanwhile, in Table \ref{tb:overall_resnet50}, the pre-perturbations do not affect QoAE obviously. On the contrary, IAE\&MI-RS performs better than other random search-based schemes in terms of QoAE because the higher query efficiency introduces fewer white noises (Fig. \ref{fig:examples-of-different-methods:e}-\ref{fig:examples-of-different-methods:h}). 

We also observe that the scores of QoAE of IAE\&MI-GE become worse than MI-GE in Table \ref{tb:overall_inceptionv3}. The reason relies on the inconsistent input sizes of Inception-V3 and VGG16. For a 299*299 clean image, we need to resize it to 224*224 in the preprocessing phase to fit the input size of VGG16. Then we also need to resize the 224*224 initial adversarial example to 299*299 in the local black-box attack phase to fit the input size of Inception-V3. The two resize operations bring extra noise. To confirm it, we resize 299*299 original images to 224*224 first and then directly resize the smaller images to 299*299. Next, we calculate MAD and PSNR between the original and the final resized images. If the two images are exactly the same, MAD and PSNR should be 0 and $+\infty$. However, the final actual MAD and PSNR are 8.52 and 37.47. If we add pre-perturbations to the smaller images, such extra noise after the second resize would be more serious. This suggests keeping the input sizes of the target model and the reference model consistent, especially when training the reference model by ourselves. 

\subsection{Impact of Parameters}
In this subsection, we discuss the effect of three parameters $\epsilon_1$, $\epsilon_2$, and $\epsilon_3$ that control the degree of perturbation. In Fig \ref{fig:examples-of-different-methods:e}-\ref{fig:examples-of-different-methods:h}, the obvious white noises damage the visual effect seriously. Thus, we first reduce $\epsilon_3$ for better image quality. Here, we test IAE\&MI-RS with different $\epsilon_3$ to attack ResNet50.

\begin{figure*}[htbp]
	\centering
	\subfigure[NoQ]{\includegraphics[scale=0.45]{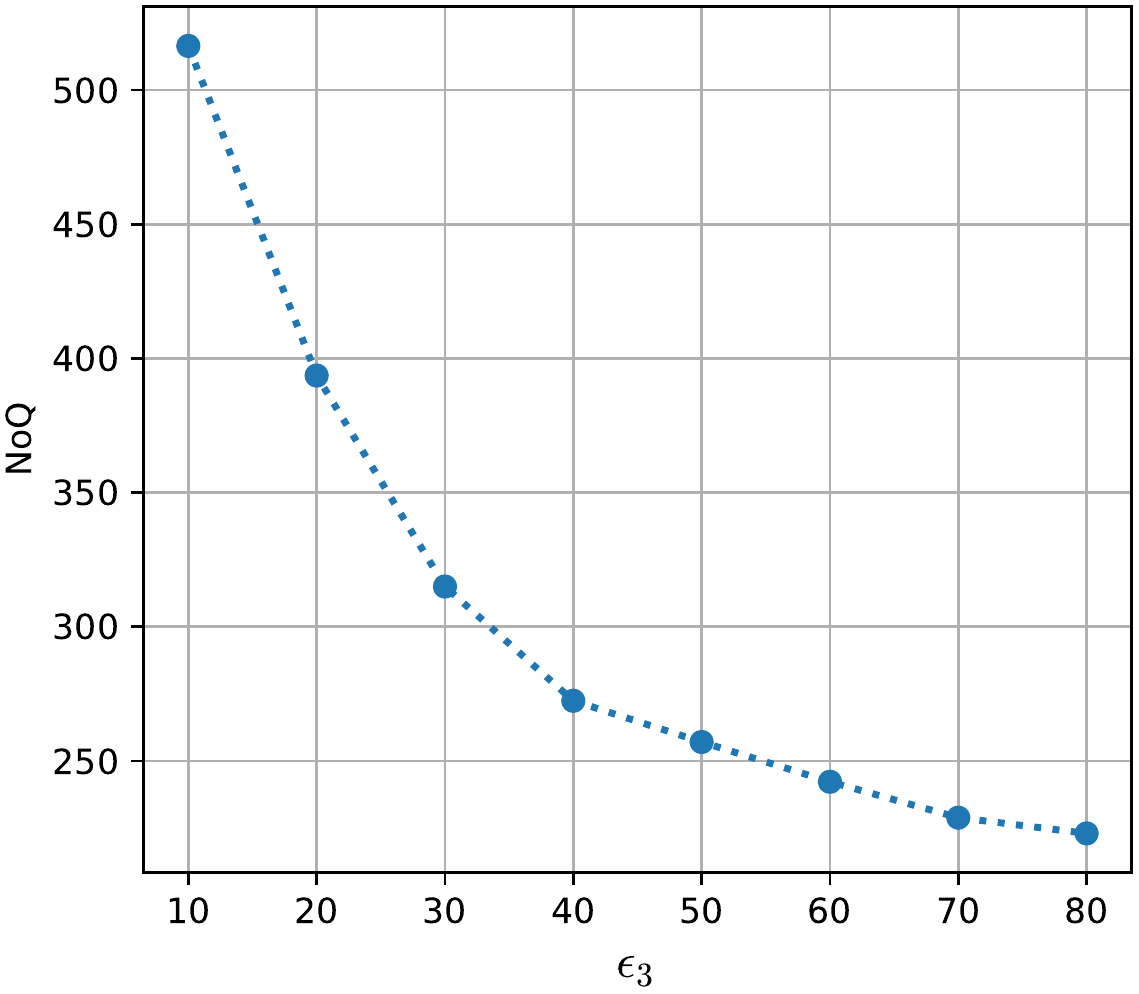}\label{fig:epsilon_3:a}}
	\subfigure[SR]{\includegraphics[scale=0.45]{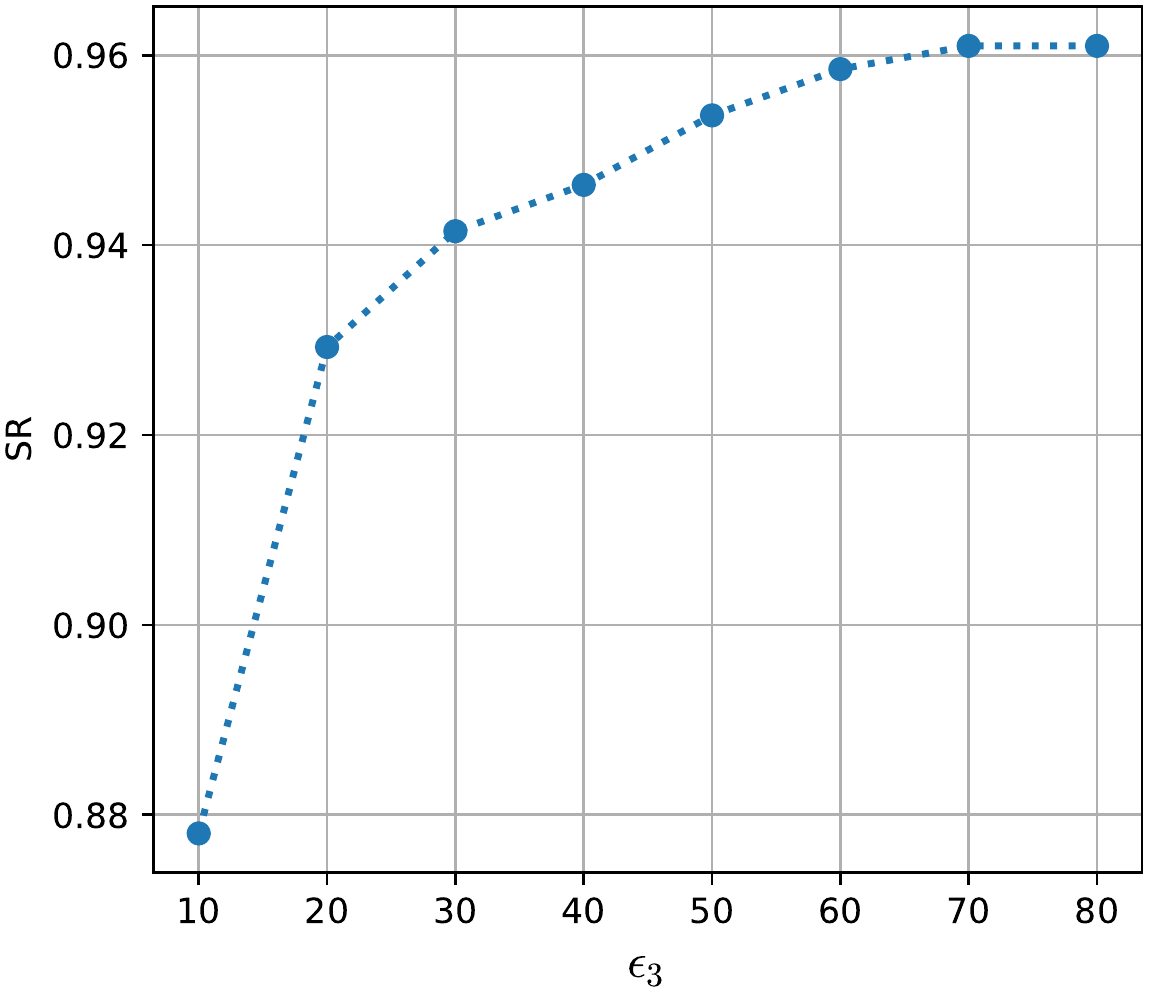}\label{fig:epsilon_3:b}}		
	\subfigure[QoAE]{\includegraphics[scale=0.45]{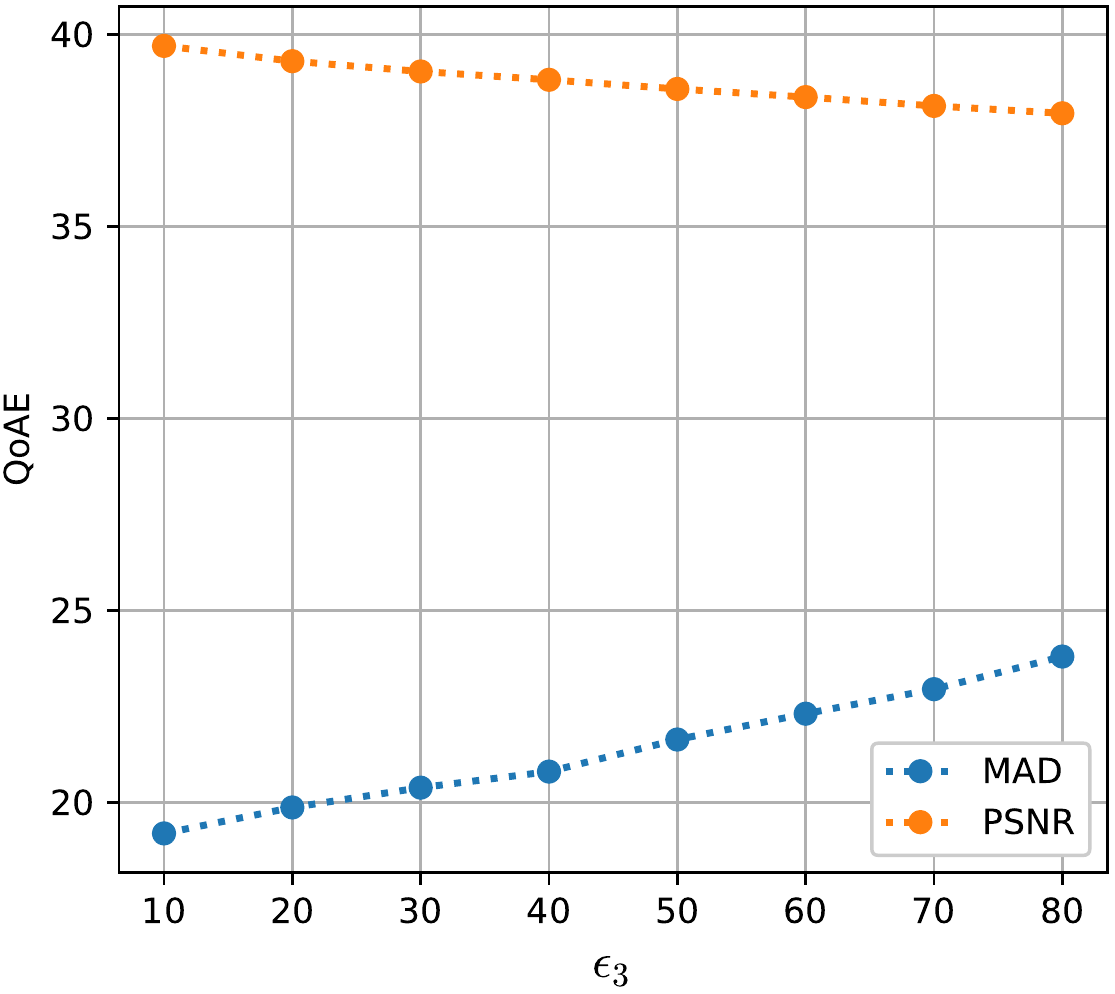}\label{fig:epsilon_3:c}}
	\caption{The effect of $\epsilon_3$ for IAE\&MI-RS on NoQ, SR and QoAE when the target model is ResNet50.}
	\label{fig:epsilon_3}
\end{figure*}

\begin{figure*}[t]
	\centering
	\subfigure[Clean]{\includegraphics[width=1.in]{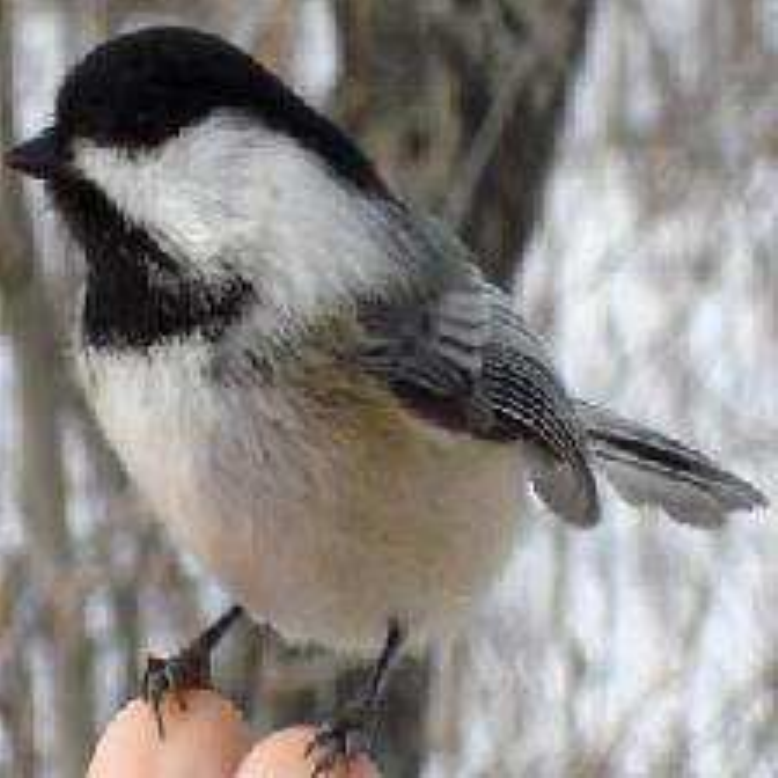}\label{fig:examples-epsi3:a}}
	\subfigure[IAE\&MI-RS(10)]{\includegraphics[width=1.in]{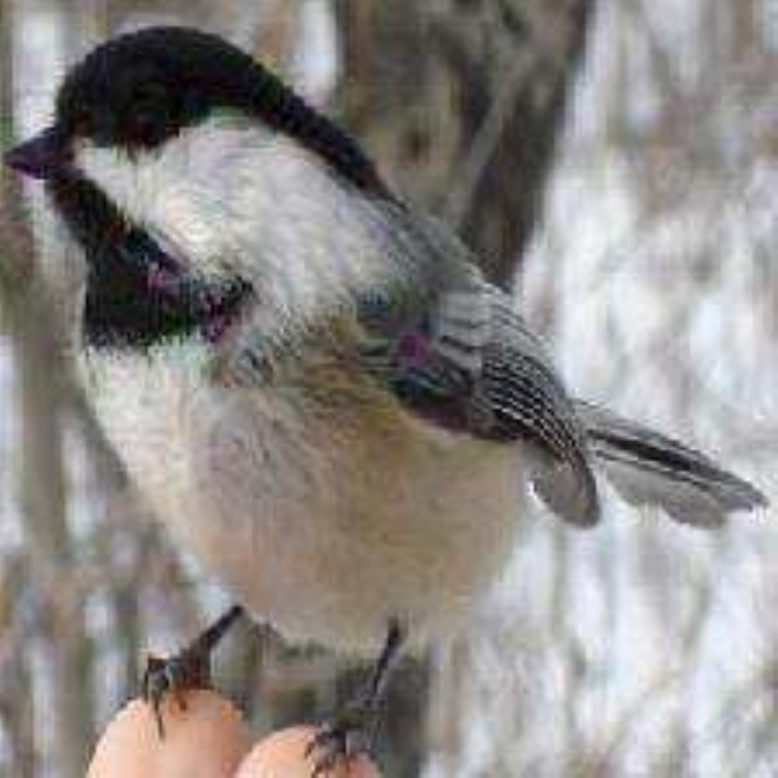}\label{fig:examples-epsi3:b}}
	\subfigure[IAE\&MI-RS(20)]{\includegraphics[width=1.in]{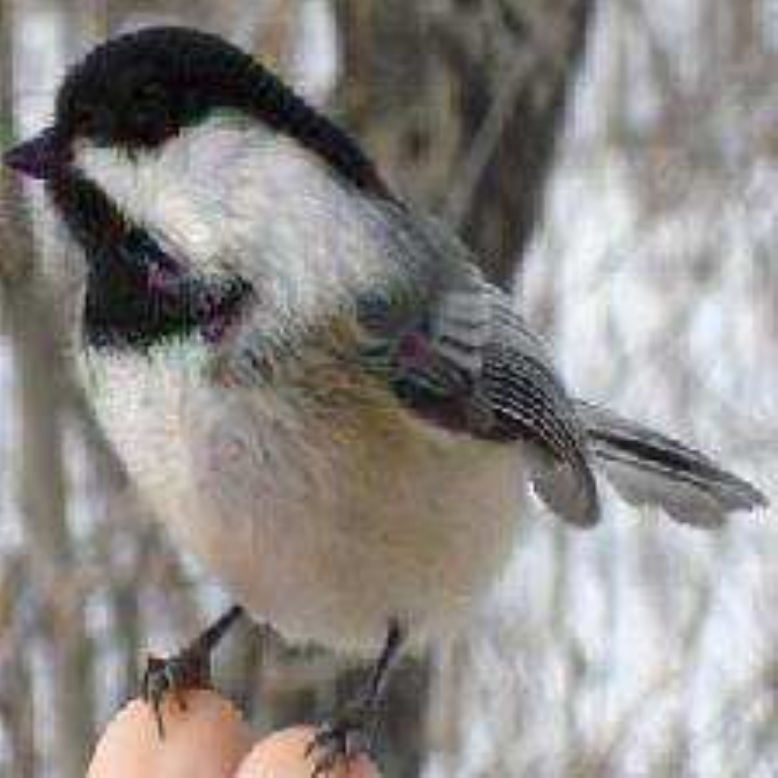}\label{fig:examples-epsi3:c}	}
	\subfigure[IAE\&MI-RS(30)]{\includegraphics[width=1.in]{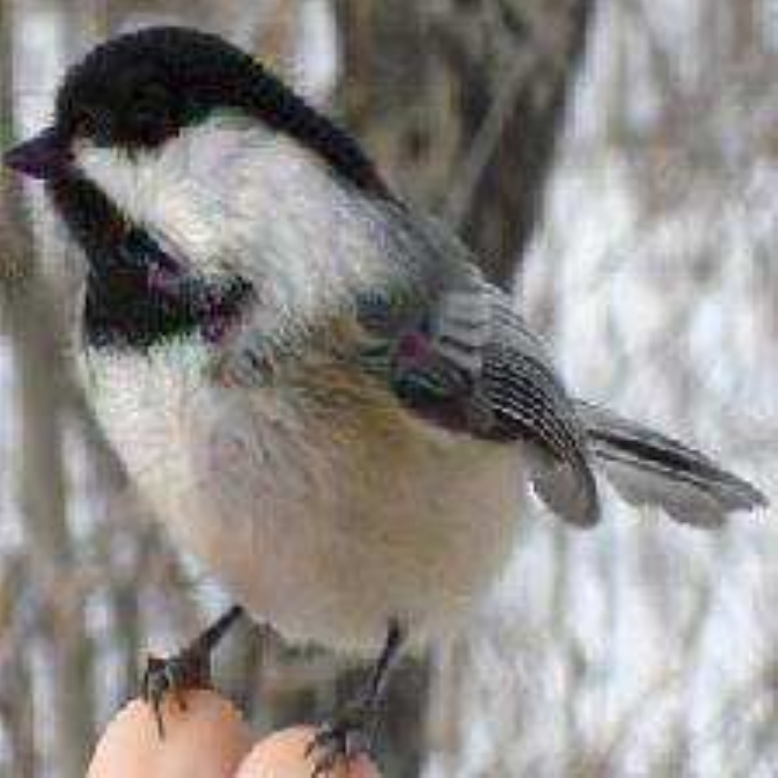}\label{fig:examples-epsi3:d}}
	\subfigure[IAE\&MI-RS(40)]{\includegraphics[width=1.in]{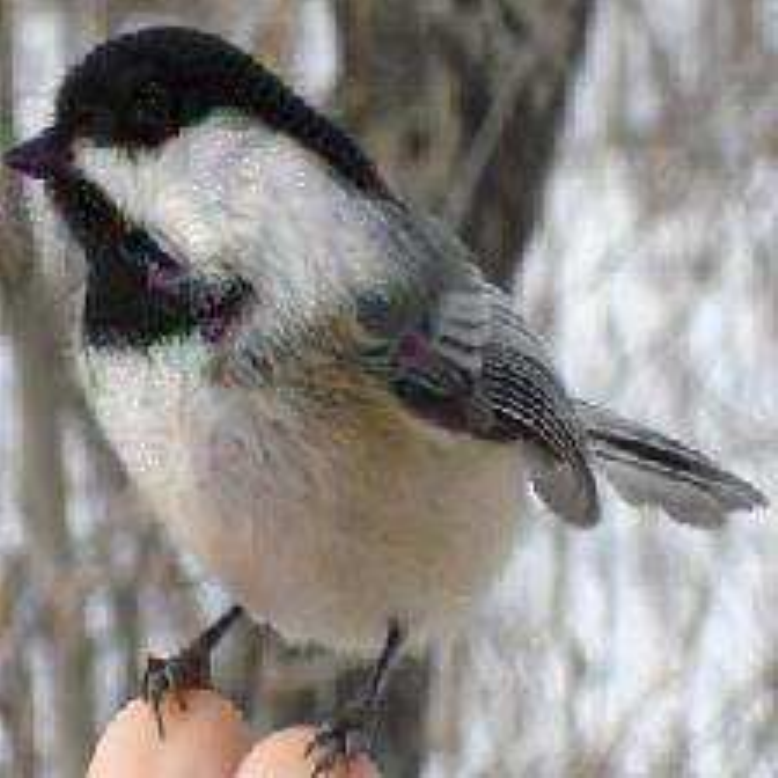}\label{fig:examples-epsi3:e}}
	\subfigure[IAE\&MI-RS(50)]{\includegraphics[width=1.in]{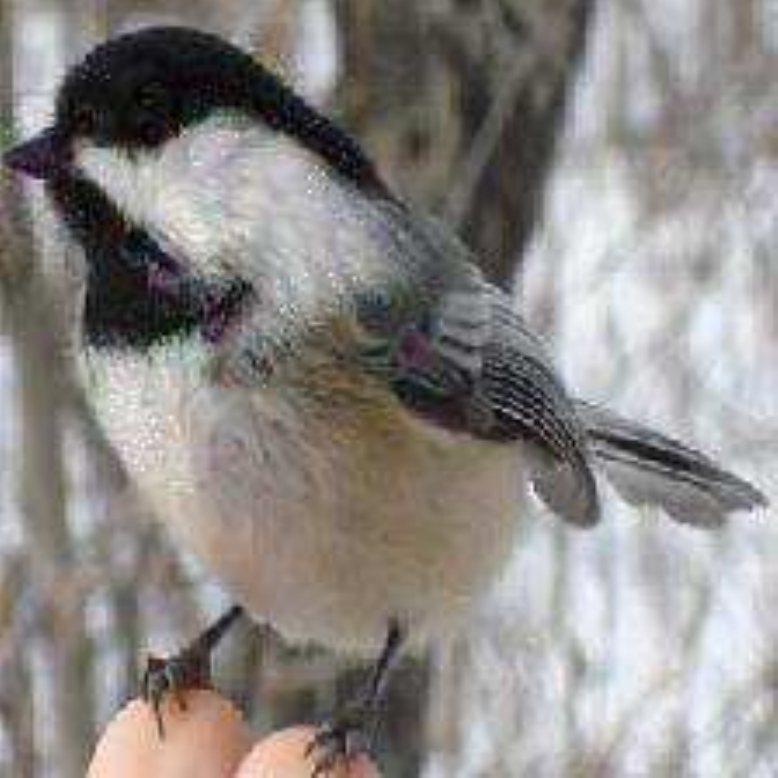}\label{fig:examples-epsi3:f}}
	
	\subfigure[IAE\&MI-RS(60)]{\includegraphics[width=1.in]{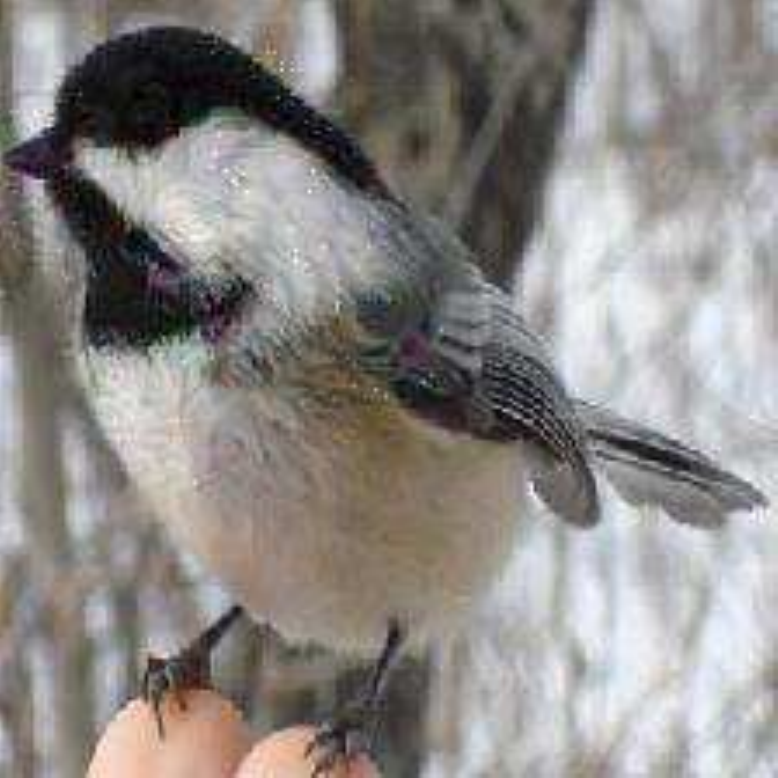}\label{fig:examples-epsi3:g}}
	\subfigure[IAE\&MI-RS(70)]{\includegraphics[width=1.in]{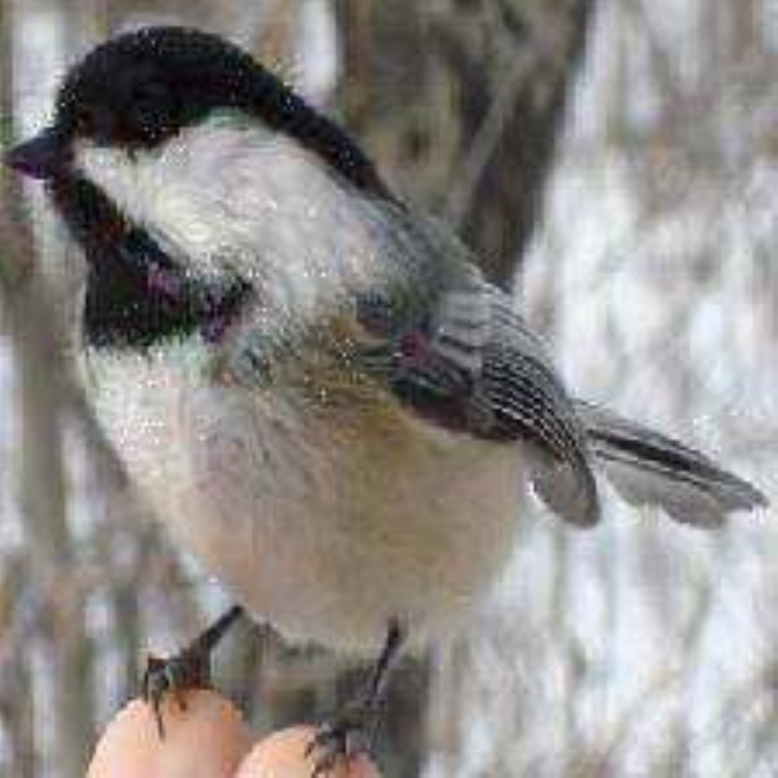}\label{fig:examples-epsi3:h}}
	\subfigure[IAE\&MI-RS(80)]{\includegraphics[width=1.in]{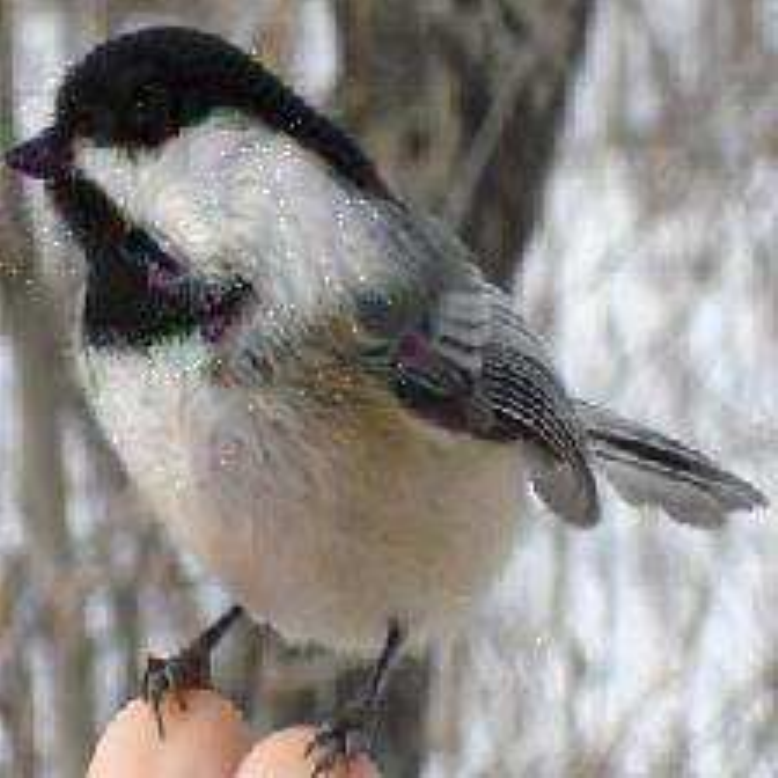}\label{fig:examples-epsi3:i}}
	\subfigure[IAE\&MI-RS(255)]{\includegraphics[width=1.in]{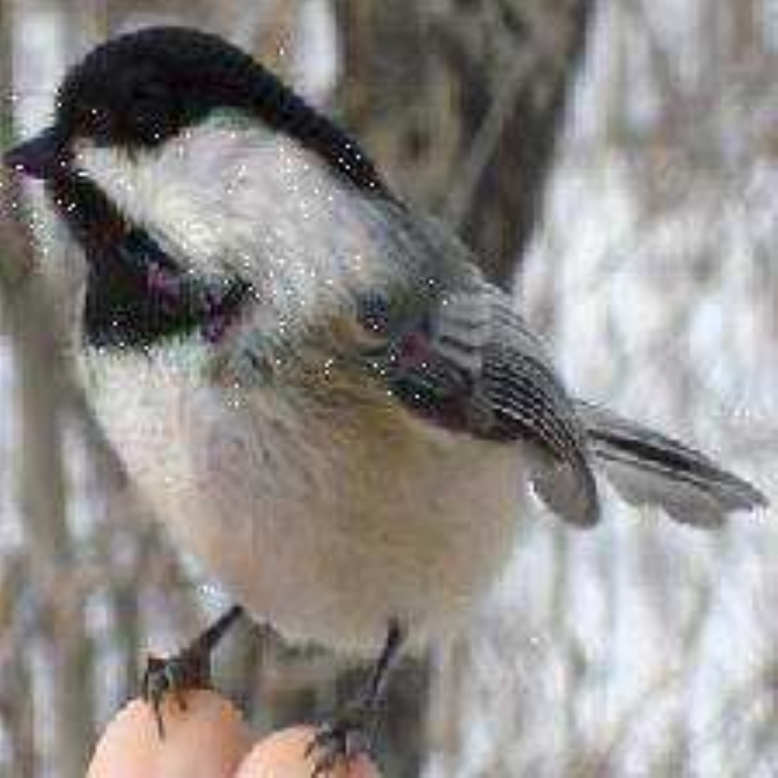}\label{fig:examples-epsi3:j}}
	\subfigure[SBLS]{\includegraphics[width=1.in]{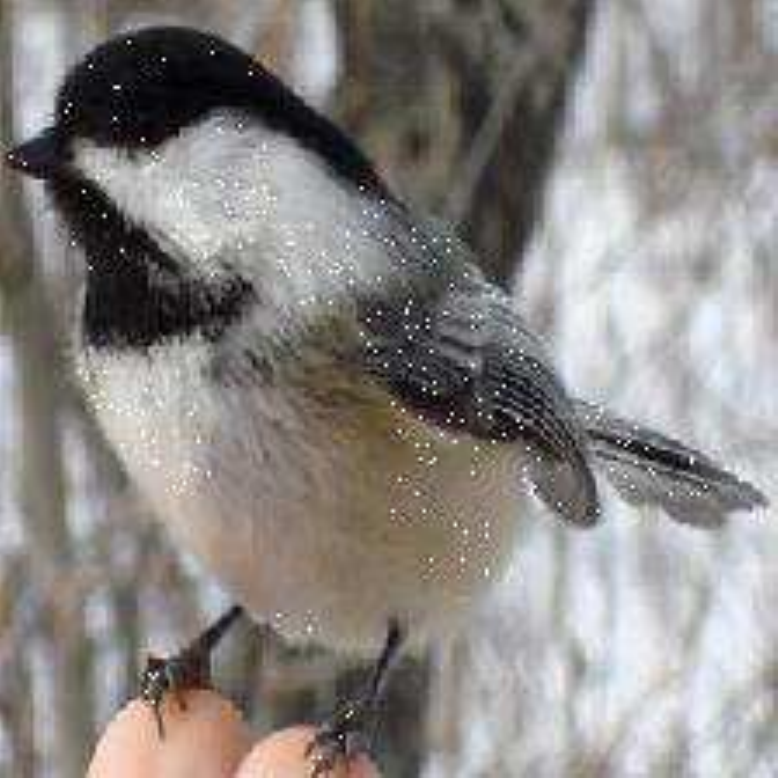}\label{fig:examples-epsi3:k}}
	\subfigure[GRS]{\includegraphics[width=1.in]{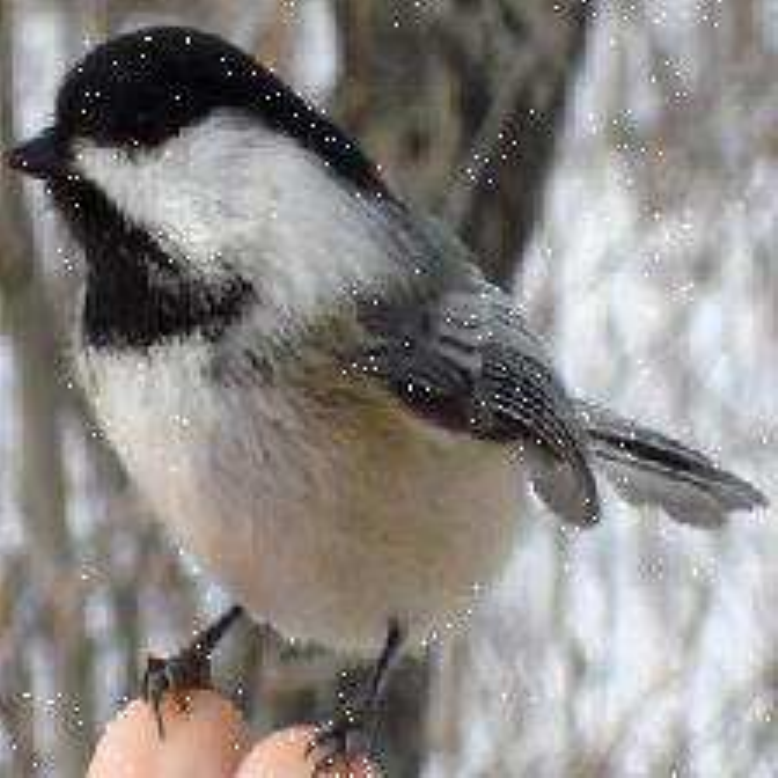}\label{fig:examples-epsi3:l}}
	\caption{Adversarial examples generated by IAE\&MI-RS, SBLS, and GRS, respectively.  (a) The original image; (b)-(j) Adversarial examples generated by IAE\&MI-RS with different $\epsilon_3$. (b) $\epsilon_3=10$, NoQ=1943, PSNR=38.42, MAD=7.96; (c) $\epsilon_3=20$, NoQ=1215, PSNR=37.57, MAD=10.23; (d) $\epsilon_3=30$, NoQ=923, PSNR=36.85, MAD=11.42; (e) $\epsilon_3=40$, NoQ=526, PSNR=36.59, MAD=12.25; (f) $\epsilon_3=50$, NoQ=515, PSNR=35.67, MAD=16.66; (g) $\epsilon_3=60$, NoQ=315, PSNR=36.06, MAD=11.27;	(h) $\epsilon_3=70$, NoQ=311, PSNR=35.42, MAD=17.85; (i) $\epsilon_3=80$, NoQ=314, PSNR=35.52, MAD=19.48; (j) $\epsilon_3=255$, NoQ=311, PSNR=31.41, MAD=39.79;(k) Adversarial example generated by SBLS and NoQ=918, PSNR=28.64, MAD=66.27; (l) Adversarial example generated by GRS and NoQ=1138, PSNR=28.34, MAD=67.42 }
	\label{fig:examples-epsi3}
\end{figure*}

In Fig. \ref{fig:epsilon_3}, we observe that a larger $\epsilon_3$ is good for reducing the NoQ and improving SR because a larger $\epsilon_3$ causes a more serious perturbation to each selected pixel. More serious perturbations mean larger offset distance such that it is more likely to achieve attacks with fewer iterations. However, a larger $\epsilon_3$ also reduces the final QoAE. In Fig. \ref{fig:examples-epsi3}, when $\epsilon_3$ is small, the perturbation produced by IAE\&MI-RS is not easy to be detected (e.g. Fig. \ref{fig:examples-epsi3:b}-\ref{fig:examples-epsi3:f}). As $\epsilon_3$ increases, the perturbation becomes more visible. 

What's more, the query efficiency of IAE\&MI-RS is better than that of GRS on average when the $\epsilon_3=20$ and is also better than that of SBLS when while $\epsilon_3=30$. Thus, IAE\&MI-RS can achieve higher query efficiency and bring much better QoAE than SBLS and GRS when using a smaller pixel perturbation.


\begin{figure*}[htbp]
	\centering
	\subfigure[NoQ]{\includegraphics[scale=0.45]{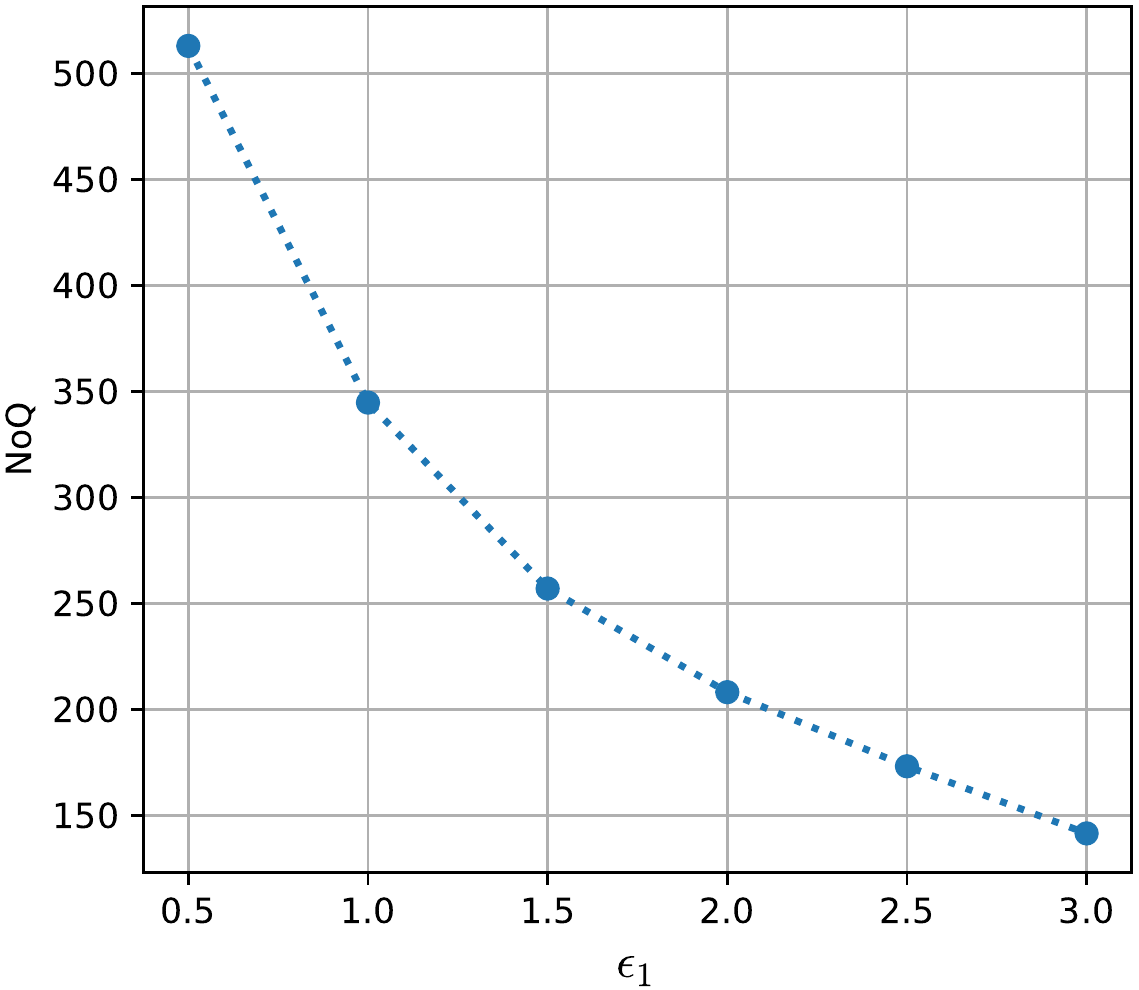}\label{fig:epsilon_1:a}}
	\subfigure[SR]{\includegraphics[scale=0.45]{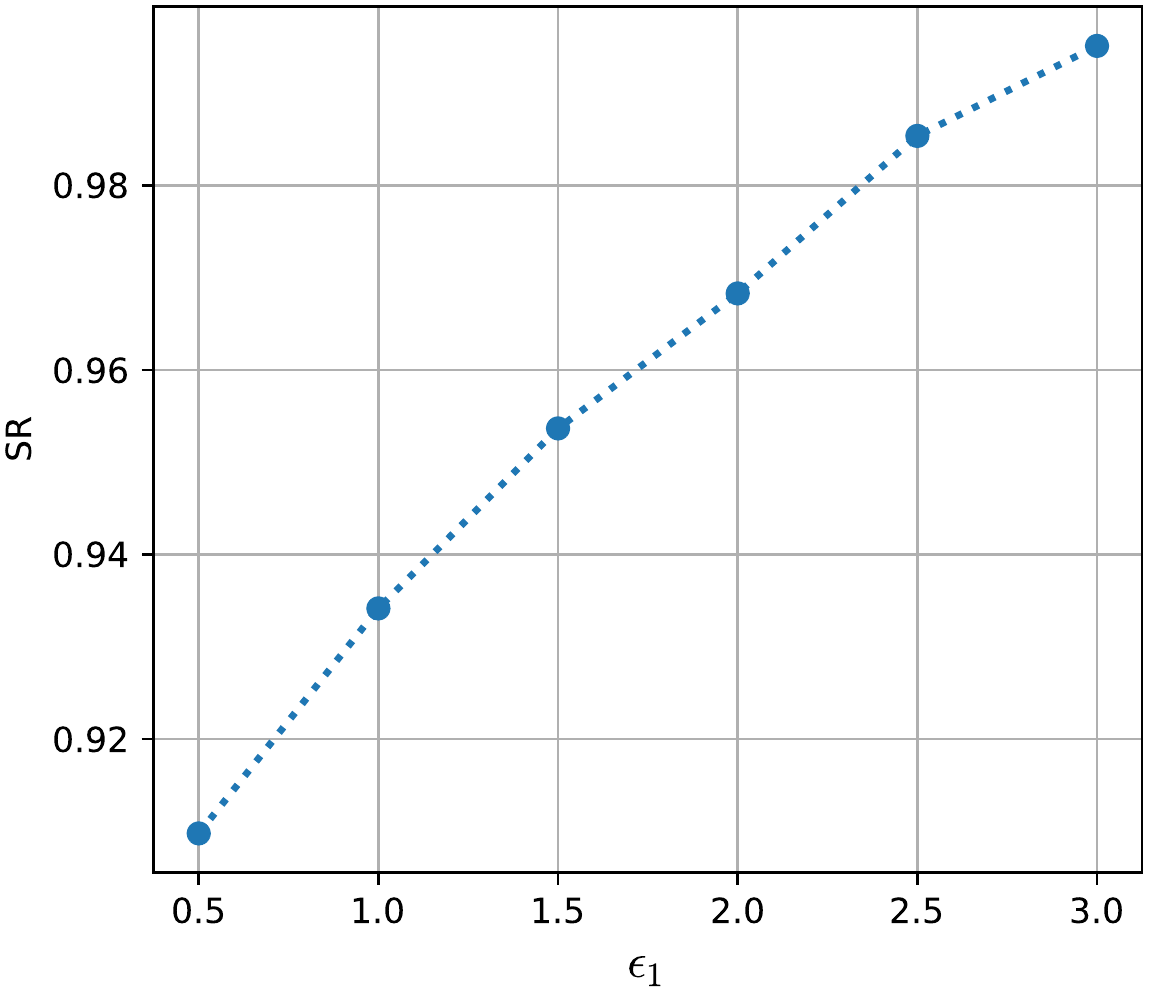}\label{fig:epsilon_1:b}}		
	\subfigure[QoAE]{\includegraphics[scale=0.45]{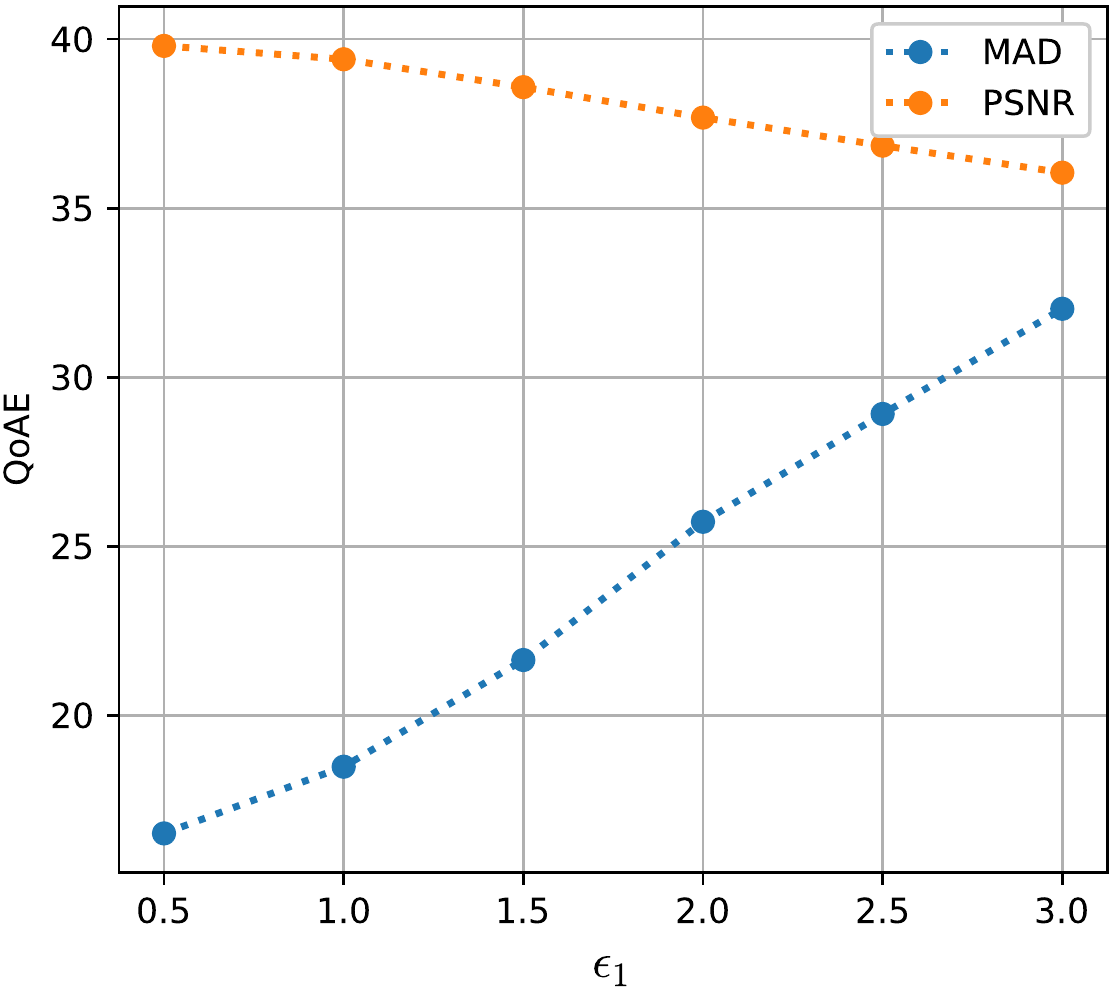}\label{fig:epsilon_1:c}}
	\caption{The effect of the $\epsilon_1$ in IAE\&MI-RS ($\epsilon_3=50$) in NoQ, SR and QoAE when the target model is ResNet50.}
	\label{fig:epsilon_1}
\end{figure*}

$\epsilon_1$ and $\epsilon_2$ also control the degree of perturbations. In theory, a larger $\epsilon_1$ or $\epsilon_2$ also improves the NoQ and the SR but reduces the QoAE just like $\epsilon_3$. Here we only further test IAE\&MI-RS with different $\epsilon_1$ that is used in the first phase to confirm it, since both $\epsilon_2$ and $\epsilon_3$ are used in the second phase just for different types of attacks. The results are shown in Fig. \ref{fig:epsilon_1} ($\epsilon_3$ is 50).

\subsection{Impact of Discriminative Areas}
In our attack framework, we introduce the transferability of model interpretations. Now, we empirically prove this phenomenon. Specifically, we show that the discriminative areas based on the transferability of model interpretations have a great effect on the target model's predictions by comparing with other types of local areas. We choose the opposite of the discriminative areas determined by Grad-CAM and denote it as NMI areas. We also choose the random areas that is called R areas for comparison. Following the way of MI-RS, we can get NMI-RS and R-RS that respectively focus on perturbing the NMI areas and the R areas.  According to the proportion of the R areas in the input, R-RS can be further denoted as R$\star$-RS, where $\star\in \left(0,1\right ]$ represents the proportion.

Besides, we introduce two cases for calculating the average of NoQ. The first one is the most common case, in which we calculate the average for all generated examples as we have stated in Section \ref{sec:experiments:criterion}. We denote it as Case-All and by default, all values are calculated in this case. The second case is considered for better reflecting the gap between two comparative schemes if there is a significant difference in their SR. In this case, we only count the adversarial examples that are successfully found by both two comparison schemes. For example, suppose there are only four clean examples: \{$\rm\textbf{x}^1$, $\rm\textbf{x}^2$, $\rm\textbf{x}^3$, $\rm\textbf{x}^4$\}. If a scheme A successfully finds adversarial examples for {$\rm\textbf{x}^1$, $\rm\textbf{x}^2$, $\rm\textbf{x}^3$} and another scheme B only finds adversarial examples for {$\rm\textbf{x}^2$, $\rm\textbf{x}^3$,$\rm\textbf{x}^4$}, when we compare A and B in terms of NoQ, we only focus on these adversarial examples generated from $\rm\textbf{x}^2$ and $\rm\textbf{x}^3$. For convenience, we denote this situation as Case-Both. We conduct experiments with the Inception-V3 as the target model. 

First, we compare MI-RS, NMI-RS, and R$\star$-RS on different input category groups in the default case (i.e. Case-All), and the results are shown in Table \ref{tb:mi_case_all}. As expected, MI-RS gets the best scores no matter in terms of NoQ or SR among all these methods in all three categories. For example, MI-RS only costs 639 queries for attack animal images and gets 85.65\% SR while other methods cost more than 800 queries and lower SR, especially the NMI-RS only achieves 61.30\%.  Such results have demonstrated the importance of the discriminative areas determined by Grad-CAM and the reference model to the target model. But it's not over. The huge gap of SR hides part of the gap of NoQ because there are many ``hard'' clean examples that can not be solved by NMI-RS or R$\star$-RS but can be solved by MI-RS. And these hard clean examples would increase the average NoQ of MI-RS in Case-All. For this reason, we re-calculate the NoQ in Case-Both.

\begin{table*}[]
	\caption{Results of MI-RS, NMI-RS and R$\star$-RS on different categories in Case-All while the target model is Inception-V3}
	\label{tb:mi_case_all}
	\centering
	\begin{tabular}{|c|c|c|c|c|c|c|c|}
		\hline
		\multicolumn{2}{|c|}{Method}                          & MI-RS                          & NMI-RS                         & R0.2-RS                        & R0.4-RS                        & R0.6-RS                        & R0.8-RS                        \\ \hline
		\multirow{2}{*}{Animal}       & NoQ                   & \textbf{639}                            & 839                            & 859                            & 819                            & 850                            & 822                            \\
		& SR                    & \textbf{85.65\%}                          & 61.30\%                          & 81.30\%                          & 80.86\%                          & 80.43\%                          & 82.17\%                          \\ \hline
		\multirow{2}{*}{Transport}    & NoQ                   & \textbf{681}                            & 917                            & 829                            & 834                            & 875                            & 814                            \\
		& SR                    & \textbf{92.15\%}                          & 69.61\%                          & 87.15\%                          & 88.24\%                          & 90.19\%                          & 88.24\%                          \\ \hline
		\multirow{2}{*}{Traffic sign} & NoQ                   & \textbf{998}                            & 1096                           & 1203                           & 1218                           & 1079                           & 1105                           \\
		& SR                    & \textbf{83.33\%}                          & 55.13\%                          & 78.20\%                          & 82.05\%                          & 75.64\%                          & 75.64\%                          \\ \hline
	\end{tabular}
\end{table*}

\begin{figure}[htbp]
	\centering
	\includegraphics[width=0.8\columnwidth]{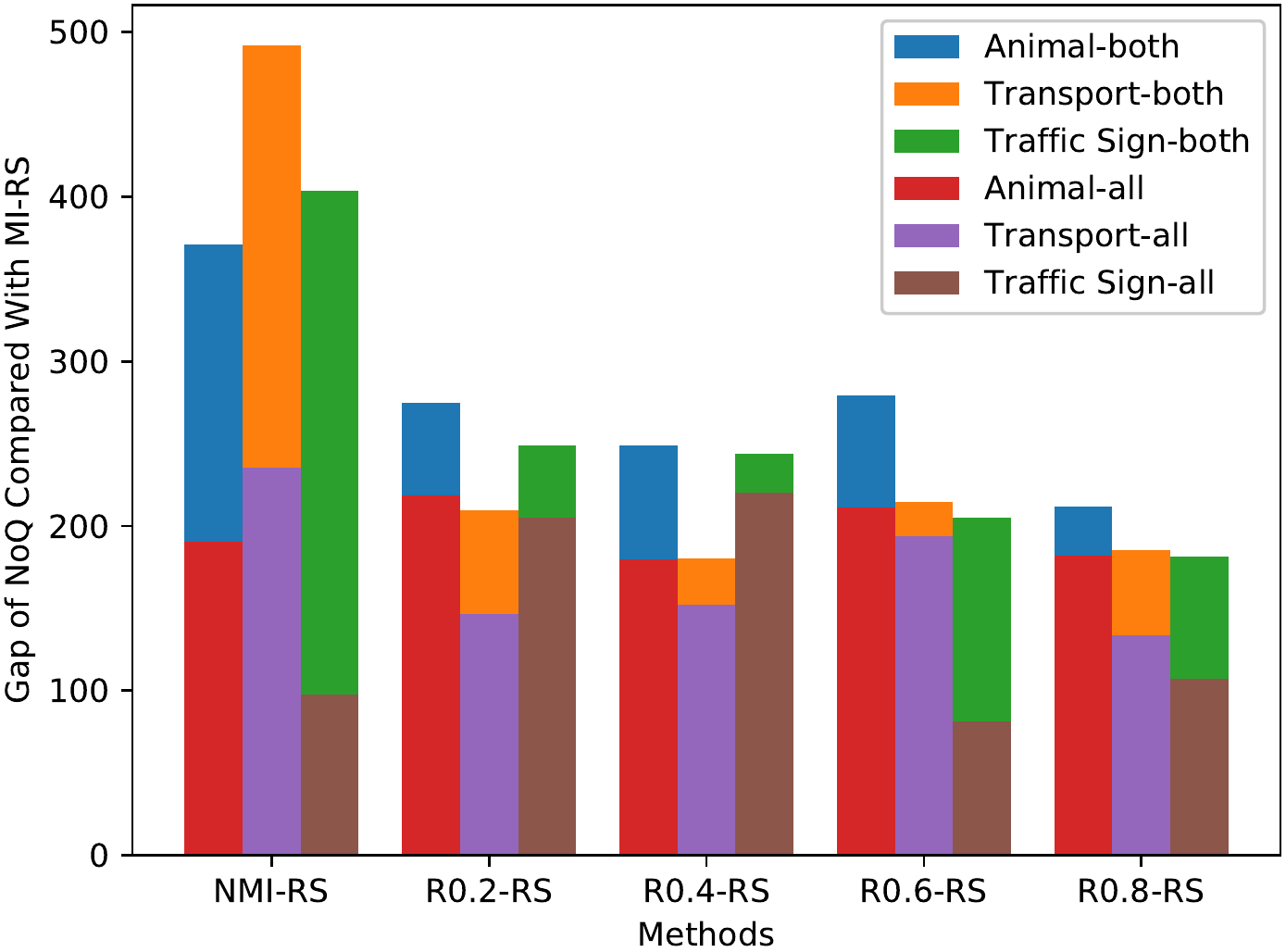}
	\caption{Gaps of NoQ between MI-RS and other comparison methods in Case-All and Case-Both. Each bar is obtained by subtracting the NoQ of MI-RS from the NoQ of the corresponding method. The ``-both" and ``-all" in the legend represents the corresponding bar is obtained in Case-Both or Case-All.}
	\label{fig:gap}
\end{figure}

\begin{table*}[]
	\caption{Results of MI-RS, NMI-RS and R$\star$-RS on different categories in Case-Both}
	\label{tb:mi_case_both}
	\centering
	\begin{tabular}{|c|c|c|c|c|c|c|c|}
		\hline
		\multirow{2}{*}{Comparative Group} & \multirow{2}{*}{Method} & \multicolumn{2}{c|}{Animal}  & \multicolumn{2}{c|}{Transport} & \multicolumn{2}{c|}{Traffic sign} \\ \cline{3-8} 
		&                         & NoQ & Examples-Both          & NoQ  & Examples-Both           & NoQ    & Examples-Both            \\ \hline
		\multirow{2}{*}{\#1}               & MI-RS                   & \textbf{412} & \multirow{2}{*}{60.43\%} & \textbf{414}  & \multirow{2}{*}{68.63\%}  & \textbf{680}    & \multirow{2}{*}{53.85\%}   \\ \cline{2-3} \cline{5-5} \cline{7-7}
		& NMI-RS                  & 783 &                        & 906  &                         & 1084   &                          \\ \hline
		\multirow{2}{*}{\#2}               & MI-RS                   & \textbf{552} & \multirow{2}{*}{79.13\%} & \textbf{606}  & \multirow{2}{*}{86.27\%}  & \textbf{945}    & \multirow{2}{*}{76.92\%}   \\ \cline{2-3} \cline{5-5} \cline{7-7}
		& R0.2-RS                 & 827 &                        & 816  &                         & 1194   &                          \\ \hline
		\multirow{2}{*}{\#3}               & MI-RS                   & \textbf{549} & \multirow{2}{*}{79.13\%} & \textbf{615}  & \multirow{2}{*}{86.27\%}  & \textbf{951}    & \multirow{2}{*}{79.49\%}   \\ \cline{2-3} \cline{5-5} \cline{7-7}
		& R0.4-RS                 & 799 &                        & 796  &                         & 1195   &                          \\ \hline
		\multirow{2}{*}{\#4}               & MI-RS                   & \textbf{545} & \multirow{2}{*}{78.26\%} &\textbf{ 638 } & \multirow{2}{*}{88.24\%}  & \textbf{869 }   & \multirow{2}{*}{74.36\%}   \\ \cline{2-3} \cline{5-5} \cline{7-7}
		& R0.6-RS                 & 824 &                        & 852  &                         & 1074   &                          \\ \hline
		\multirow{2}{*}{\#5}               & MI-RS                   & \textbf{571} & \multirow{2}{*}{80.00\%} & \textbf{629}  & \multirow{2}{*}{88.24\%}  & \textbf{914}    & \multirow{2}{*}{74.36\%}   \\ \cline{2-3} \cline{5-5} \cline{7-7}
		& R0.8-RS                 & 783 &                        & 814  &                         & 1096   &                          \\ \hline
	\end{tabular}
\end{table*}
Specifically, we compare MI-RS with each other methods and get six pairs of comparative methods. In Case-Both, we only care about the adversarial examples that are found by both two methods for the same clean examples. We set Examples-Both to represent such clean examples. The results in Case-Both are presented in Table \ref{tb:mi_case_both}, where the percentage of Examples-Both is the proportion of the size of Examples-Both to the number of examples in the corresponding category in our dataset. We can see that all these percentages of Examples-Both in Table \ref{tb:mi_case_both} are very close to the corresponding SR values in Table \ref{tb:mi_case_all}. For example, The SR of NMI-RS on animals in Table \ref{tb:mi_case_all} is 61.30\%, and in Table \ref{tb:mi_case_both} the percentage of Examples-Both for MI-RS and NMI-RS on animals is 60.43\%.  Meanwhile,  we can find that the gap of NoQ between MI-RS and other compared methods grows larger in Case-Both. To be intuitive, we have drawn Fig. \ref{fig:gap}, which reflects the gap of NoQ in Case-All and Case Both. In Fig. \ref{fig:gap}, the bar calculated in Case-Both is higher than that calculated in Case-All. Thus, if NMI-RS or R$\star$-RS can find an adversarial example for one input, MI-RS can not only find the corresponding adversarial example with a high probability but also be more query-efficient.

\subsection{Impact of Reference Model}
Last but not least, we discuss the impact of the reference model on the performance of our attacks, especially on the NoQ and SR. All previous works that use auxiliary models assume the model is well-trained. But sometimes we may not be able to get such a good model, especially when we train it by ourselves with limited training data. Thus, it is necessary to analyze the impact of the performance of the auxiliary model on the attack, which is missing in most previous works. Here we explore a more practical scenario, in which we only have a reference model with limited performance.

We carry out an experiment on Dogs vs. Cats task \footnote{https://www.kaggle.com/c/dogs-vs-cats} that only needs to classify dogs and cats. We assume that there is no publicly available reference model for the target task but we can train one through transfer learning. Transfer learning allows us to avoid training a model from scratch and has been wildly adopted in model training. We also assume that the amount of training data is limited. We confirm that even the model performance is influenced by the amount of training data, the reference model can be still used to improve the query efficiency and SR in our methods compared with global attacks.

The official website provides 25000 training samples. We randomly choose 1000 samples for testing model performance and remove them from the original training set. After that, a target model TM is trained based on the all remained 24000 training samples and data augmentation for the best accuracy possible. In the experiments, we construct TM based on VGG19 by replacing the fully-connected layers of VGG19 with new and randomly initialized fully-connected layers to adapt to the new task. During training, we freeze the weights in all convolution layers and only train the parameters in the new fully-connected layers. We construct the reference model based on VGG16 and train it with different numbers of training samples. As result, we get four reference models: RM\_50, RM\_100, RM\_1000, and RM\_5000 while the numbers 50, 100, 1000, 5000 represent the number of training samples of each category. All training samples are resized to 150*150 and are mapped into the range of [0,1]. 

The accuracy of the TM and the four reference models is shown in Table \ref{tb:acc}. We observe that the limited training samples influence the model performance. As the number of training samples reduces, the model accuracy becomes lower and lower. Next, we randomly choose 100 clean examples (50 cats and 50 dogs) and use IAE\&MI-RS to attack the TM with the four reference models respectively. Besides, we also use GRS to attack the TM for comparison. 
\begin{table}[htbp]
	\caption{Accuracy of the target model and all reference models}
	\label{tb:acc}
	\centering
	\begin{tabular}{|c|c|c|c|c|c|}
		\hline
		Model&TM  & RM\_50&RM\_100& RM\_1000& RM\_5000 \\ \hline
		Accuracy&\textbf{92.2\%}&57.4\%  &77.2\%&86.4\%&88.6\%\\ \hline
	\end{tabular}
\end{table}

Experimental results are illustrated in Table \ref{tb:dogvscat1}. In general, as the reference model performance increase, the attack effect of IAE\&MI-RS becomes better. We can observe that even we use RM\_100 as the reference model that only acquires 77.2\% accuracy, IAE\&MI-RS improves the query efficiency by 54.16\% and loses only 1\% SR compared with GRS. When we use RM\_1000 as the reference model, the improvement of NoQ is more significant without reducing the SR compared with GRS. Thus, we conclude that our framework can be compatible with reference models with lower performance. On the other hand, a model with very poor performance, like the RM\_50, is not suitable as the reference model. We can observe that RM\_50 brings the lowest SR that only be 79\% even the NoQ is 19 less than that of GRS. The huge SR gap makes IAE\&MI-RS non-attractive.

\begin{table}[htbp]
	\caption{NoQ and SR of GRS and IAE\&MI-RS with different reference models}
	\label{tb:dogvscat1}
	\centering
	\begin{tabular}{|c|c|c|c|c|c|}
		\hline
		\multirow{2}{*}{Method} & \multicolumn{4}{c|}{IAE\&MI-RS}        & \multirow{2}{*}{GRS} \\ \cline{2-5}
		& RM\_50 & RM\_100 & RM\_1000 & RM\_5000 &                      \\ \hline
		NoQ                     & 426    & 204     & 106      & \textbf{71}       & 445                  \\ \hline
		SR                      & 79\%     & 98\%     & 99\%      & \textbf{100\%}      & 99\%                  \\ \hline
	\end{tabular}
\end{table}

    \section{Discussion} \label{sec:discussion}
\subsection{Defenses}

In this paper, we focus on designing query efficient local adversarial attacks with the black-box access of the targeted model. From the defense aspect, a number of defense schemes have been proposed to obtain a more robust model, or detect or remove adversarial examples. Here, we analyze the effect of existing state-of-the-art defenses as follows.

Adversarial training \cite{goodfellow2015explaining} is an efficient defense that retrains the targeted model with adversarial examples to improve the model's robustness. It can be adopted to defend our attacks if the adversarial attack techniques for generating the pre-perturbed and final examples are known to the defender.
Otherwise, as shown in \cite{li2019adversarial}, it would be costly to collecting all potential adversarial examples because of the low transferability of adversarial training. For example, a model that is retrained for defending an adversarial attack may still fail to defend another new attack.

The defender can also adopt other defenses \cite{li2019adversarial} (e.g. adversarial example detection, randomization, rounding confidence scores, and noise filters). It is an arms race between adversarial attacks and defenses and the conclusions in \cite{li2019adversarial} can be applied in our attacks because our attacks are also local like that in \cite{li2019adversarial}.

\subsection{Multiple Reference Models}
We can also adopt multiple reference models to further improve the query efficiency as in \cite{Yanpei2017delving,suya2020hybrid}. In this case, we can get more gains because the stronger transferability of adversarial examples makes the initial adversarial example have a larger probability to directly fool the target model. We only need to change the details of the first phase for adapting to multiple reference models. First, we respectively apply Grad-CAM to the reference models, and the union of the white areas of all binary maps determines the final discriminative areas. The union ensures that the final areas are likely to contain the complete local areas the target model depends on. Then we produce the initial adversarial example through the reference models as follows:
\begin{equation}
	{\rm{\textbf{x}}}_{t+1}=clip\left({\rm{\textbf{x}}}_{t}+\epsilon_1*(sign(\nabla_{{\rm{\textbf{x}}}_{t}}H)\otimes sign(UBM)) \right)
\end{equation}
where
\begin{align}
	&H=\sum_{i}^{U}J({\rm{\textbf{x}}}_{t},y;\theta_{{f_r}_i})\\
	&UBM=\sum_i^U BM_i
\end{align}
$U$ is the number of reference models. $\theta_{{f_r}_i}$ is the parameters of the $i$-th reference model and $BM_i$ is the binary map produced through the $i$-th reference model. We can also produce the initial adversarial example using the optimization in \cite{Yanpei2017delving,suya2020hybrid} instead of directly using the direction of the gradients. This approach is applicable when there are multiple reference models and the computational resources are sufficient.

\subsection{Impact of Model Interpreters}
Besides Grad-CAM, other state-of-the-art model interpreters can be also applied to our framework, including Grad-CAM++ \cite{8354201} and Smooth Grad-CAM++ \cite{omeiza2019smooth}. The key insight of choosing model interpreters is to select a model interpreter that can accurately identify the discriminative areas of clean examples. The families of class activation mapping-based interpreters \cite{8354201,omeiza2019smooth,zhou2016learning,selvaraju2017grad} have been proven to be effective to identify the discriminative areas in white-box attacks \cite{9157746} and without loss of generality, we use Grad-CAM in our experiments.

    \section{Conclusion} \label{sec:conclusion}
In this paper, we explore how to launch local black-box attacks with limited query overhead. We only perturb the discriminative areas of clean examples because they have a greater impact on model's predictions. To determine the discriminative areas, we adopt model interpretations according to our key observation on the transferability of model interpretations. This property allows us to identify the discriminative areas through a reference model rather than a more complex semantic segmentation model \cite{li2019adversarial}. What's more, we can also produce pre-perturbations through the same reference model due to the transferability of adversarial examples, which significantly improves the query efficiency. Finally, based on the two properties, we propose a novel black-box, query-efficiency framework to generate local perturbations. Experimental results show that our framework can reduce the query overhead and bring better visual effect and attack success rate, even the performance of the reference model is limited.

    \bibliographystyle{IEEEtran}
    \bibliography{main}

\end{document}